**Sensing environmental physical interaction to traverse cluttered obstacles**

Yaqing Wang, Ling Xu, Chen Li*

Department of Mechanical Engineering, Johns Hopkins University

*Corresponding author: [redacted]

## RUNNING TITLE

SENSING FORCES TO TRAVERSE CLUTTERED OBSTACLES

## NOVELTY STATEMENT

Robots often rely on computer vision to map a geometry map to avoid obstacles, but they cannot traverse cluttered obstacles using this approach, hindering many applications. Potential energy landscapes have been recently established as an approach to guide robots in controlling physical interactions to traverse cluttered obstacles. However, they cannot be measured by vision. Here, we demonstrated that robots can sense obstacle contact forces to estimate potential energy landscapes.

## ABSTRACT

The long-standing, dominant approach to robotic obstacle negotiation relies on mapping environmental geometry to avoid obstacles. However, this approach does not allow for traversal of cluttered obstacles, hindering applications such as search and rescue operations through earthquake rubble and exploration across lunar and Martian rocks. To overcome this challenge, robots must further sense and utilize environmental physical interactions to control themselves to traverse obstacles. Recently, a physics-based approach has been established towards this vision. Self-propelled robots interacting with obstacles results in a potential energy landscape. On this landscape, to traverse obstacles, a robot must escape from certain landscape basins that attract it into failure modes, to reach other basins that lead to successful modes.

Thus, sensing the potential energy landscape is crucial. Here, we developed new methods and performed systematic experiments to demonstrate that the potential energy landscape can be estimated by sensing environmental physical interaction. We developed a minimalistic robot capable of sensing obstacle contact forces and torques for systematic experiments over a wide range of parameter space. Surprisingly, although these forces and torques are not fully conservative, they match the potential energy landscape gradients that are conservative forces and torques, enabling an accurate estimation of the potential energy landscape. Additionally, a bio-inspired strategy further enhanced estimation accuracy. Our results provided a foundation for further refining these methods for use in free-locomoting robots. Our study is a key step in establishing a new physics-based approach for robots to traverse clustered obstacles to advance their mobility in complex, real-world environments.

## 1 INTRODUCTION

Sensing one's surrounding environment helps animals and robots move through the complex world. Animals combine vision and other sensory modalities (mechanosensing, chemosensing, etc.) to sense, plan, and control their physical interaction with obstacles to *traverse* them (Sober and Sabes, 2005; Wessnitzer and Webb, 2006; Wehner, 2020). By contrast, robots mostly rely on visual information to navigate complex environments while *avoiding* obstacles (Koditschek, 2021) (**Fig. 1**, top row). Specifically, a robot often uses long-range visual sensing (e.g., cameras, LiDAR) (e.g., **Fig. 1A**) to create a geometric map of the environment (e.g., **Fig. 1B**). Using this map, it constructs an artificial potential field (Khatib, 1986; Rimon and Koditschek, 1992), with its goal modeled by a global minimum and obstacles modeled by high potential regions (**Fig. 1C**). It then plans and follows a gradient descent path towards the goal around the obstacles (Tsitsiklis, Bertsekas and Athans, 1986), often transitioning between various locomotor modes (e.g., **Fig. 1D**).

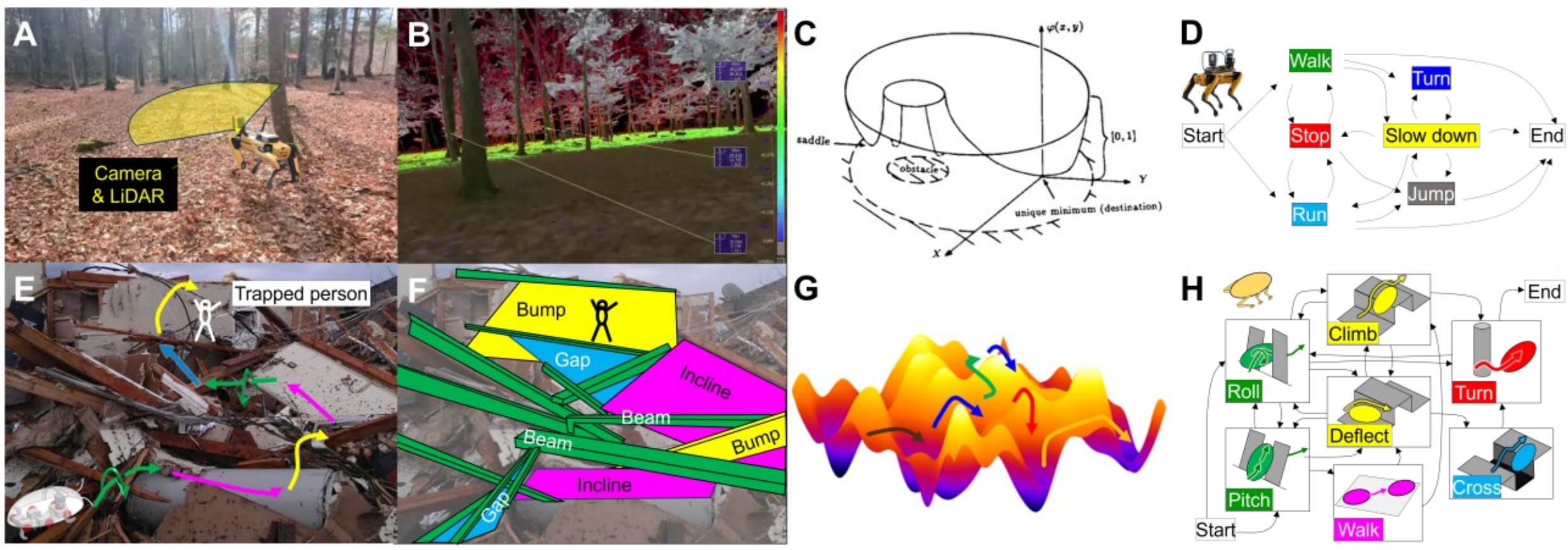

**Fig. 1. Envisioned potential energy landscape approach enabling legged obstacle traversal analogous to artificial potential field enabling obstacle avoidance.** (**A**) A legged robot leverages vision-based sensing to navigate a flat, sparse forest terrain with scattered trees. (**B**) Geometric map scanned using LiDAR from the robot in (A). (**C**) Artificial potential field approach for obstacle avoidance. (**D**) Multi-pathway locomotor transitions to avoid obstacles. (**E**) Envisioned traversal of cluttered, large obstacles (e.g., earthquake rubble for search and rescue). (**F**) Examples of abstracted large obstacles that present diverse locomotor challenges. Robots must sense their environmental interaction physics with such obstacles to enable robust traversal. (**G**) Potential energy landscape approach for obstacle traversal. (**H**) Multi-pathway locomotor transitions enable traversal via sensing and controlling physical interactions with obstacles. (A, B) Image courtesy of Intuitive Robots (https://youtu.be/s3zWXRLaAFw). (C–H) Adapted from (Rimon and Koditschek, 1992; Othayoth, Thoms and Li, 2020; Othayoth *et al.*, 2021).

However, this vision-based obstacle avoidance approach to robot navigation usually fails in environments with densely cluttered (spacing ~ body size), large (~ body size) obstacles (Clifton *et al.*, 2023), because a collision-free trajectory to the goal may simply not exist (Kalakrishnan *et al.*, 2010; Lee *et al.*, 2020). To traverse, a robot must physically interact with obstacles to generate appropriate forces and torques. However, a lack of sensing physical interaction with obstacles leads to poor control of locomotion through them. This has diminished the usefulness of robots in many important applications, such as search and rescue in rubble after earthquakes and hurricanes (Murphy, 2014) (**Fig. 1E**), environmental monitoring in mountain boulders and forest debris (Angelini *et al.*, 2023), and planetary exploration through large Martian and Lunar rocks (Li and Lewis, 2023).

To traverse densely cluttered terrains, a robot should sense not only environmental geometry but also physical interaction (obstacle contact forces and torques) and use this information to appropriately control its self-propulsion (Othayoth, Thoms and Li, 2020; Othayoth *et al.*, 2021). Towards this vision, our lab recently established a new approach to modeling the environmental physical interaction of robots (and animals) using potential energy landscapes (**Fig. 1**, bottom), which applies to a diversity of large obstacles that present distinct locomotor challenges (e.g., **Fig. 1E, F**) (Li *et al.*, 2015; Gart and Li, 2018; Gart *et al.*, 2018; Othayoth, Thoms and Li, 2020; Han *et al.*, 2021; Othayoth *et al.*, 2021; Wang, Othayoth and Li, 2022). Analogous to artificially defined potential fields for geometry-based obstacle avoidance, these potential energy landscapes from first principle (**Fig. 1G**) provide a physics-based foundation to compose locomotor transitions (**Fig. 1H**) for obstacle traversal (for a review, see (Othayoth *et al.*, 2021)).

Specifically, a self-propelled robot's or animal's body physically interacts with obstacles, resulting in a potential energy landscape (the system's real potential energy as a function of body 3-D position and rotation), with attractive basins of stability separated by potential energy barriers (**Fig. 1G**). Due to continual self-propulsion breaking continuous frictional contacts, the system's state tends to settle into these basins. As the robot or animal is attracted to each stability basin on the landscape, its motion emerges as a distinct locomotor mode (which often involves large body rotations, not just translation as in obstacle avoidance) (**Fig. 1H**). Given the limited propulsive forces and torques of the robot or animal, some modes lead to traversal, which are more favorable to the robot or animal, while others lead to being trapped, which are less favorable. Thus, to traverse obstacles, the robot or animal must propel to destabilize itself from attraction by the less favorable basins/modes of entrapment and overcome potential energy barriers to transition to the more favorable basins/modes that lead to traversal. Because these barrier-crossing transitions are highly strenuous (requiring the generation of large propulsion and work compared to the robot's or animal's capacity (Othayoth *et al.*, 2021; Spence, Wilshin and Byrnes, 2022)), it is beneficial to estimate the potential energy landscape using sensory information and achieve a trajectory to traverse with the least resistance and effort. This approach can enable robots to traverse densely cluttered obstacles that would otherwise be impossible due to the limited propulsion. However, we had not solved the problems of how robots can sense environmental physical interaction and estimate the potential energy landscapes of unknown obstacles.

We hypothesized that robots can estimate the potential energy landscape by sensing obstacle contact forces and torques. In a robot–obstacle interaction system, the negative gradients of the potential energy landscape are conservative forces and torques, which contribute to obstacle contact forces and torques (see **Sec. S.3** for derivation). However, obstacle contact forces and torques also have contributions from non-conservative forces (frictional force, damping force, inertial force, etc.) and torques. Yet, our lab's previous works demonstrated that the system's locomotor mode transition dynamics are strongly governed by the potential energy landscape in densely cluttered terrains (Othayoth *et al.*, 2021). This suggests that the conservative forces and torques (landscape gradients) dominate and the non-conservative forces and torques are small. In other words, the measured obstacle contact forces and torques (including both conservative and non-conservative components) may provide a good estimate of the conservative forces and torques (i.e., potential energy landscape gradients).

Testing this hypothesis requires creating a robot with custom onboard sensors to measure obstacle contact forces and torques around its body, because the diverse locomotor modes useful for traversing densely cluttered obstacles involve large body rotations (Othayoth *et al.*, 2021), which cause various parts of the body surface to contact obstacles. For sensing environmental contact forces and torques, robots often rely on proprioceptive sensing by either using commercial force and torque sensors (e.g., from ATI, OnRobot, FUTEK) (Qinghua Li, Takanishi and Kato, no date; Hirai *et al.*, 1988), measuring the deformation elastic components (Grizzle *et al.*, 2009), or inferring from motor torques (Bledt *et al.*, 2018). Thanks to recent advances, robots may use exteroceptive sensing, including using flexible sensory arrays (Sundaram *et al.*, 2019; Kamegawa *et al.*, 2020; Yao *et al.*, 2020; Zhu *et al.*, 2020; Roberts, Zadan and Majidi, 2021; Ramesh, Fu and Li, 2022). However, self-propulsion during interaction with large, cluttered obstacles always induces continual collisions with the obstacles (Othayoth *et al.*, 2021), for which sophisticated commercial sensors are too fragile and expensive to have multiple ones. Flexible sensory arrays usually only provide a single normal force instead of 3-D forces (Ha, Lim and Ko, 2018; Yao *et al.*, 2020; Zhu *et al.*, 2020; Roberts, Zadan and Majidi, 2021), have high latency (due to sequential analog reading of a voltage on each unit (Luo *et al.*, 2017; Lee *et al.*, 2019)), and nonlinearity (e.g., from creep behavior (Cholleti *et al.*, 2021; Ramesh, Fu and Li, 2022)). These challenges motivated us to develop custom sensors.

Here, we take the next step towards our vision by creating a robot capable of sensing obstacle contact forces and torques distributed on it and testing how well the potential energy landscape can be estimated from the sensed forces and torques over a wide range of the parameter space. Specifically, the robot body was instrumented with several custom, low-cost 3-D force sensors and many custom contact sensors distributed over its surface. They enabled the robot to measure both 3-D obstacle contact forces and positions, thereby determining the resulting torques. To test our hypothesis, we controlled the robot to propel forward through cluttered obstacles in prescribed trajectories with a broad range of body rotations, and we tested how well the sensed obstacle contact forces and torques matched the conservative forces and torques (i.e., potential energy landscape gradients) and how well the potential energy landscape was estimated. We further studied whether a bio-inspired strategy improved the estimation accuracy.

## 2 RESULTS

### 2.1 Model system to guide experimental design

Our study focused on a model system of grass-like beam obstacle traversal, building on our lab's previous works (Li *et al.*, 2015; Othayoth, Thoms and Li, 2020; Wang, Othayoth and Li, 2022) (**Fig. 2A, B**). To traverse stiff, cluttered beam obstacles with gaps narrower than its body width, the discoid cockroach or a cockroach-inspired robot often transitions from a less favorable, strenuous pitch mode (pushing forward across beams with large body pitching), which requires a large propulsive force and mechanical energy cost (**Fig. 2A, B**, blue), to a more favorable, much easier roll mode (rolling into beam gaps and maneuvering through), which requires a much smaller propulsion and mechanical energy cost (**Fig. 2A, B**, red). When the trajectory of the system **(Fig. 2C)** is viewed on the potential energy landscape over the animal's or robot's body roll–pitch space (Euler angle follows the Tait-Bryan convention) **(Fig. 2D)**, this pitch-to-roll transition (**Fig. 2C, iii'**) requires escaping entrapment in a pitch basin (**Fig. 2D, ii, iii**, blue), crossing a potential energy barrier, and reaching a roll basin (**Fig. 2D, ii, iii**, red). These basins emerge and morph as the body moves forward, changing the pitch-to-roll transition potential energy barrier (**Fig. 2D, ii, iii**, gray dashed curve).

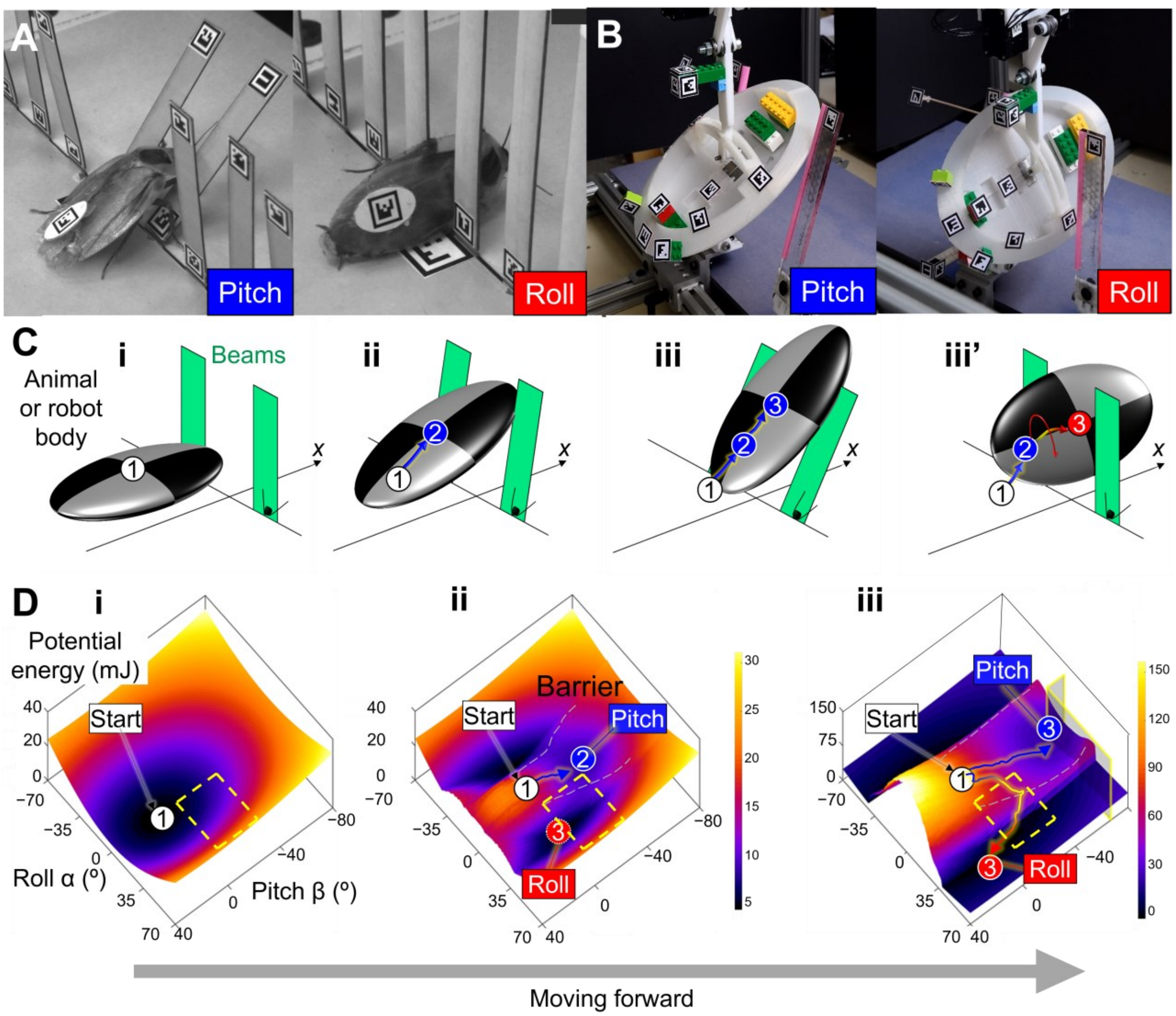

**Fig. 2. Model system of grass-like cluttered beam obstacle traversal.** (**A, B**) When (A) a cockroach and (B) a minimalistic, feedforward robot traverse cluttered grass-like beams with small gaps (< body width), they either use a pitch mode (left, blue) or a roll mode (right, red) (Othayoth, Thoms and Li, 2020). (**C**) Schematics of an animal or robot's body interacting with two beams. (i–iii or iii') System configuration when the body is not in contact with the beams (i), when the body interacts with beams in the pitch mode (ii, iii), or when the body transitions to the roll mode (iii'). (**D**) Snapshots of the potential energy landscape over roll-pitch (α-β) space. Before physically interacting with the beams (i), the landscape has a global basin. During interaction (ii, iii), a pitch basin and left- and right-roll basins emerge on the landscape, separated by potential energy barriers (gray dashed curves). The pitch or roll mode (blue or red circles in (C)) emerges as the system is attracted to the pitch or roll basin (blue or red circle in (D)), respectively. Gray dashed curves on the landscapes show potential energy barriers. Arrows on the landscapes show examples of state trajectory. Yellow dashed boxes show the roll–pitch space of the landscape estimated in **Sec. 2.4**. Adapted from (Othayoth, Thoms and Li, 2020).

Based on these observations, here we focused on studying whether we can create a robot capable of sensing obstacle contact forces and torques during interaction with cluttered beams, and how well we can use them to estimate the potential energy landscape over the body roll–pitch space in a region around the saddle point of the potential energy barrier separating the pitch and roll basins, as the body moves forward across the beams (**Fig. 2D,** yellow dashed boxes).

The previous work also observed that the cockroaches repeatedly flexed their heads and rubbed their pronotums against the obstacle edges when traversing cluttered obstacles (**Fig. 3**). We speculate that this is an active sensing behavior, by applying sensor motion to enhance environmental sensing (Okada and Toh, 2006; Baba, Tsukada and Comer, 2010; Schütz and Dürr, 2011; Van Breugel and Dickinson, 2014; Bush, Solla and Hartmann, 2016; Bajcsy, Aloimonos and Tsotsos, 2018; Berendes and Dürr, 2022). Specifically, the back-and-forth oscillation might result in a pair of frictions of roughly the same amplitude and opposite direction, which cancel each other out in a time average.

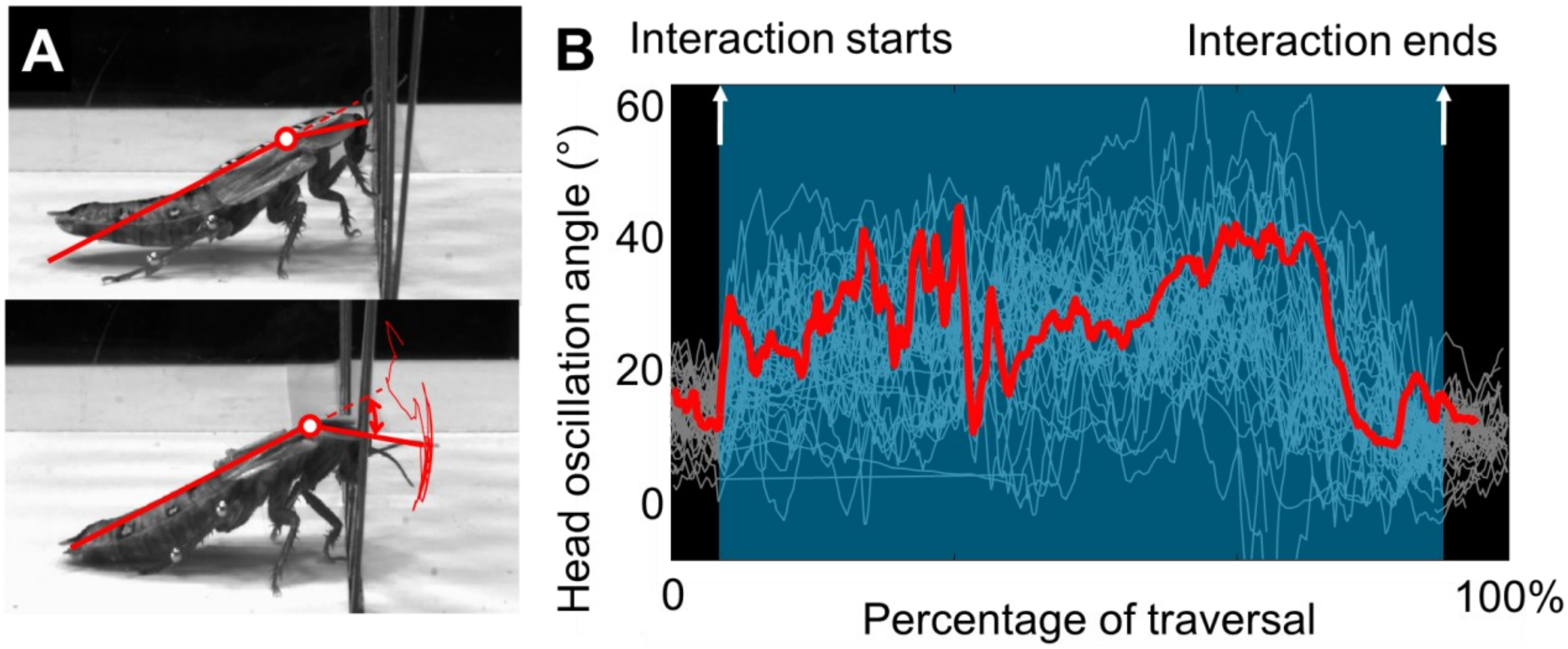

**Fig. 3. Cockroaches repeatedly flexed their heads as a bio-inspired sensor motion.** (**A**) Snapshots of a cockroach repeatedly flexing its head while traversing cluttered, large obstacles. Head flexion angle was the relative pitch between the head and thorax. Red curve shows the representative head flexion angle from a trial. (**B**) Head flexion angle as a function of traversal progress. The traversal progress was defined as in (Wang, Othayoth and Li, 2022) (without going into detail, it normalizes time across trials of different times for the animal to traverse). Thick red curve is from the representative trial shown in (A). White curves show 35 other trials. Cyan shadow shows where the animal physically interacted with obstacles. Adapted from Figure 7 of (Wang, Othayoth and Li, 2022).

Because friction and its resulting torque are a major non-conservative component of the contact forces and torques, cancellation by itself may reduce the difference between the contact forces and torques and the conservative ones. Inspired by the animal, a robot may benefit from similar oscillatory sensor motions for friction cancellation, improving landscape estimation. To test this, we conducted another set of experiments in which we further separated the robot's head from its main body and repeatedly flexed the head, mimicking animal behavior. We studied whether this improved the estimation of the potential energy landscape gradients.

## 2.2 New robot capable of obstacle contact force and torque sensing

We created a minimalistic, cockroach-inspired robot (**Fig. 4A**) that can sense obstacle contact forces and torques (see **Sec. S.1** for details). We followed the previous sensor-less robot to study the passive dynamics of locomotor-obstacle interaction (Othayoth, Thoms and Li, 2020), which consisted of a simple body with an ellipsoid-like shape similar to the cockroach's and a pair of deflectable beams (**Fig. 2B**). Similarly, the new robot body was cropped from the same ellipsoid-like shape (**Fig. 4B**, translucent green). To enable a similar oscillatory head flexion as the animals', we used motors to connect the robot's head to the main body. To avoid being caught by the beams and manage wires (see **Sec. S.2** for rationale), we further removed the outer shell of the main body. The beams were rigid plates attached to the bases with Hookean torsional springs, the same as in the previous study (Othayoth, Thoms and Li, 2020). The rotational joints of the beams were preloaded so that the beams were initially vertical.

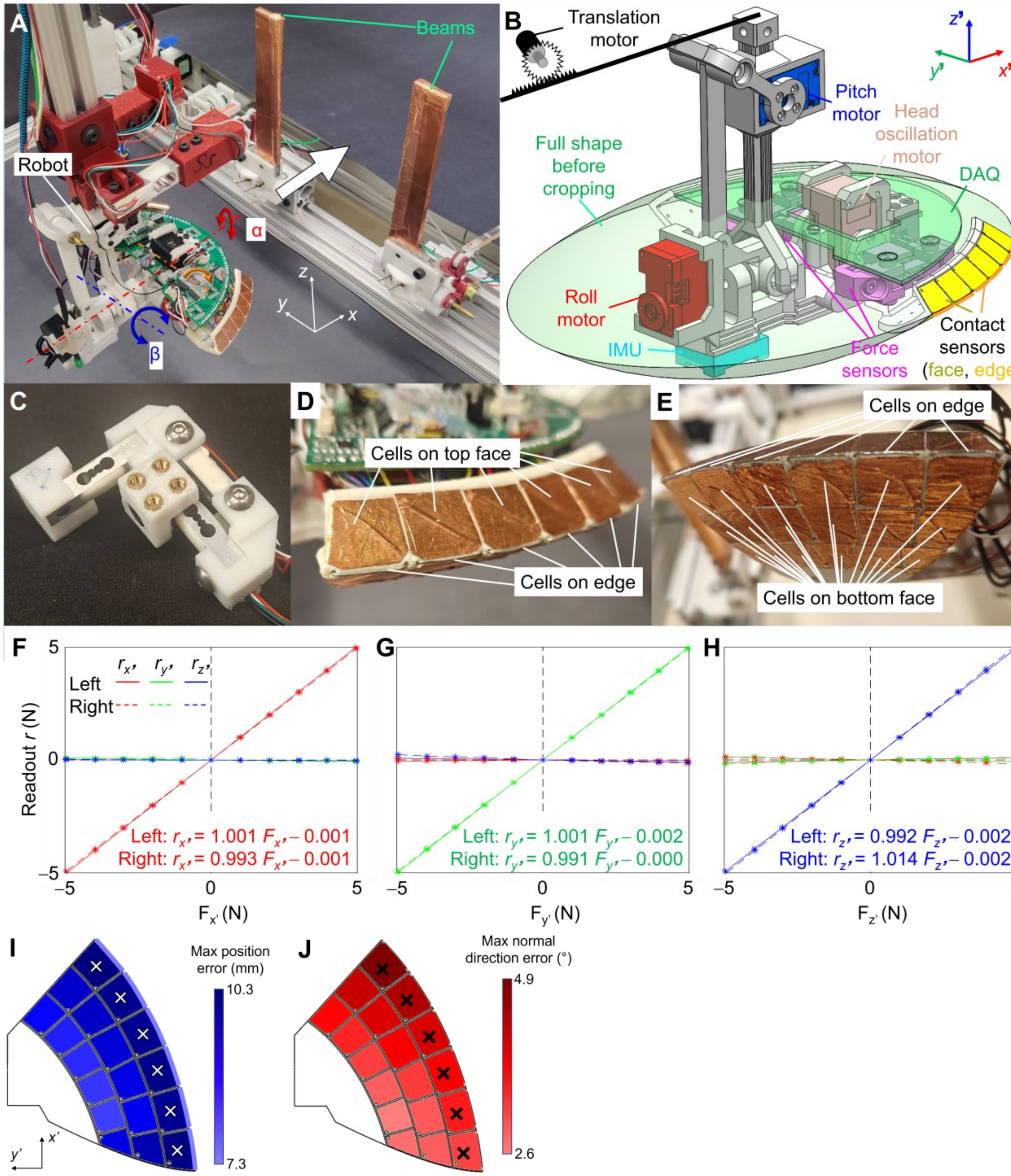

**Fig. 4. New robot for sensing obstacle contact forces and torques.** (**A**) Photo of the system consisting of a robot body with force and contact sensors and two beams. *x*, *y*, and *z* axes show the lab frame. White, red, and blue arrows show translation in fore-aft (*x*) and rotation in roll (α) and pitch (β) directions, respectively. (**B**) CAD model of the body. The shells were cropped from an ellipsoid-like shape (translucent green). The body is propelled forward along +$x$ direction (white arrow in (A)) by a servo motor via a gear-rack mechanism (black). Body roll and pitch (red and blue arrows in (A)) are controlled by servo motors (red and blue). Head pitch relative to the body is controlled by another servo motor (pink). Magenta shows custom 3-D force sensors. Yellow and orange show contact sensors on the face and the edge, respectively. Cyan shows an inertial measurement unit (IMU, cyan). Green shows a DAQ board. $x$'-, $y$'-, and $z$'- axes show body frame. (**C**) Custom 3-D force sensor. (**D, E**) Contact sensors on the faces and edges. (D) Oblique top view. (E) Oblique bottom view. (**F–H**) Force sensor calibration results. Curves show the sensor readout $r$ (red: $x$', green: $y$', blue: $z$') as a function of applied force $F$ in body (F) $x$'-, (G) $y$'-, and (H) $z$'-axes, respectively. Solid and dashed curves are from the left and right sensors, respectively (note that it is hard to differentiate the curves from left and right sensors because the two sensors are highly identical). Formulas show results from linear regression of sensor readout $r$ on applied force $F$ in the same degrees of freedom. (**I, J**) Maximal (I) position and (J) normal direction errors from each contact sensor. The pattern of contact sensors is from the bottom view. Because the left and right shells are mirrored, only the right shell is presented. Contact sensors on the top surface have the same pattern as those marked with crosses. $x$'- and $y$'-axes show body frame. Color on each contact sensor shows the error magnitude. Note that the normal direction error of those on the edge was not calculated. See **Sec. 4.1** and **Sec. S.7** for definitions.

To sense obstacle contact forces and torques, we added force and contact sensors to the head (**Fig. 4C-E**). Specifically, to measure the contact force with either beam, we separated the head into an inner core and two outer shells (left and right). Between either shell and the core, we added a custom, small (58 mm × 44 mm × 22 mm), low-cost 3-axis force sensor (**Fig. 4C**) consisting of load cells to measure 3-D contact forces. Calibration experiments (**Sec. 4.1**) comparing the applied and measured forces (**Fig. 4F-H**) showed that the custom force sensors had a nearly perfect precision along all three degrees of freedom ($R^2 = 0.999$) and small crosstalk in different degrees of freedom (< 4%).

To obtain 3-D contact torques, we added 60 pieces of contact sensors distributed over the outer surface of the head to detect the contact position on the head with either beam (**Fig. 4D, E**). In particular, we sectioned the faces and edges of the head. We covered each section on the faces with a piece of conductive copper tape (**Fig. 4B,** yellow) and each section on the edges with a piece of copper wire (**Fig. 4B,** orange). Each piece of copper tape or wire was connected to the DAQ (**Fig. 4B,** green), which formed a contact sensor. We also covered the beam surfaces with grounded copper tape so that contacts could be detected (**Sec. 4.1**). The contact position was estimated at the center of the contacted sensor. We calculated the resulting torques using the measured force and contact position (**Sec. 4.3**). The sectioning pattern of the contact sensors was carefully designed (**Sec. 4.1** and **Sec. S.7**) so that the estimation errors of contact position and normal direction (used when subtracting friction in **Sec. S.7**) at contact points were small (i.e., < 11 mm, < 5°, respectively, see **Sec. 4.1** and **Sec. S.7** for definitions) (**Fig. 4I, J**).

The robot body was tethered and driven forward (**Fig. 4A**, white) by a motor via a gear-rack mechanism (**Fig. 4B,** black) to emulate the forward propulsion generated by legs, but without adding legs. To apply a broad range of body roll and pitch for estimating the potential energy landscape over the body pitch–roll space, we added two motors (**Fig. 4B,** red and blue) to control body roll and pitch (instead of a gyroscope mechanism that allowed free pitching and rolling in the previous study (Othayoth, Thoms and Li, 2020)).

To enable oscillatory head flexion similar to that of a cockroach's head (Wang, Othayoth and Li, 2022), the head was actuated by a motor (**Fig. 4B**, pink) to further pitch around an axis in the front of the body (like the neck joint of a cockroach). When oscillatory head flexion was enabled, the head flexed to an amplitude of 20° at 2 Hz, which emulated the animal behavior after scaling (see **Sec. S.6** for frequency selection). See **Sec. S.2** for the iterative development of the robotic system.

### 2.3 Measured obstacle contact forces and torques match potential energy landscape gradients

To test the hypothesis that the robot can use obstacle contact forces and torques to estimate the potential energy landscape, we systematically varied the fore-aft position $x$ and roll $\alpha$ and pitch $\beta$ of the robot body. Before each trial, the body roll and pitch were set to the desired values, and then it was driven forward to traverse the beam obstacles, while the force sensors measured the 3-D obstacle contact forces $\overrightarrow{F_1}$ and $\overrightarrow{F_2}$ (**Fig. 5A, B**, red) and the contact sensors measured the contact positions on the body $\overrightarrow{r_1}$ and $\overrightarrow{r_2}$ with both beams (**Fig. 5A, B**, orange) as a function of the robot body's fore-aft position $x$ at 50 Hz. From these, we calculated the total fore-aft force $\overrightarrow{F_x}$ and roll torque $\overrightarrow{T_\alpha}$ and pitch torque $\overrightarrow{T_\beta}$ about the body center (defined as the center of the ellipsoid in **Fig. 4B**) as a function of fore-aft position $x$ (**Fig. 5C-F**).

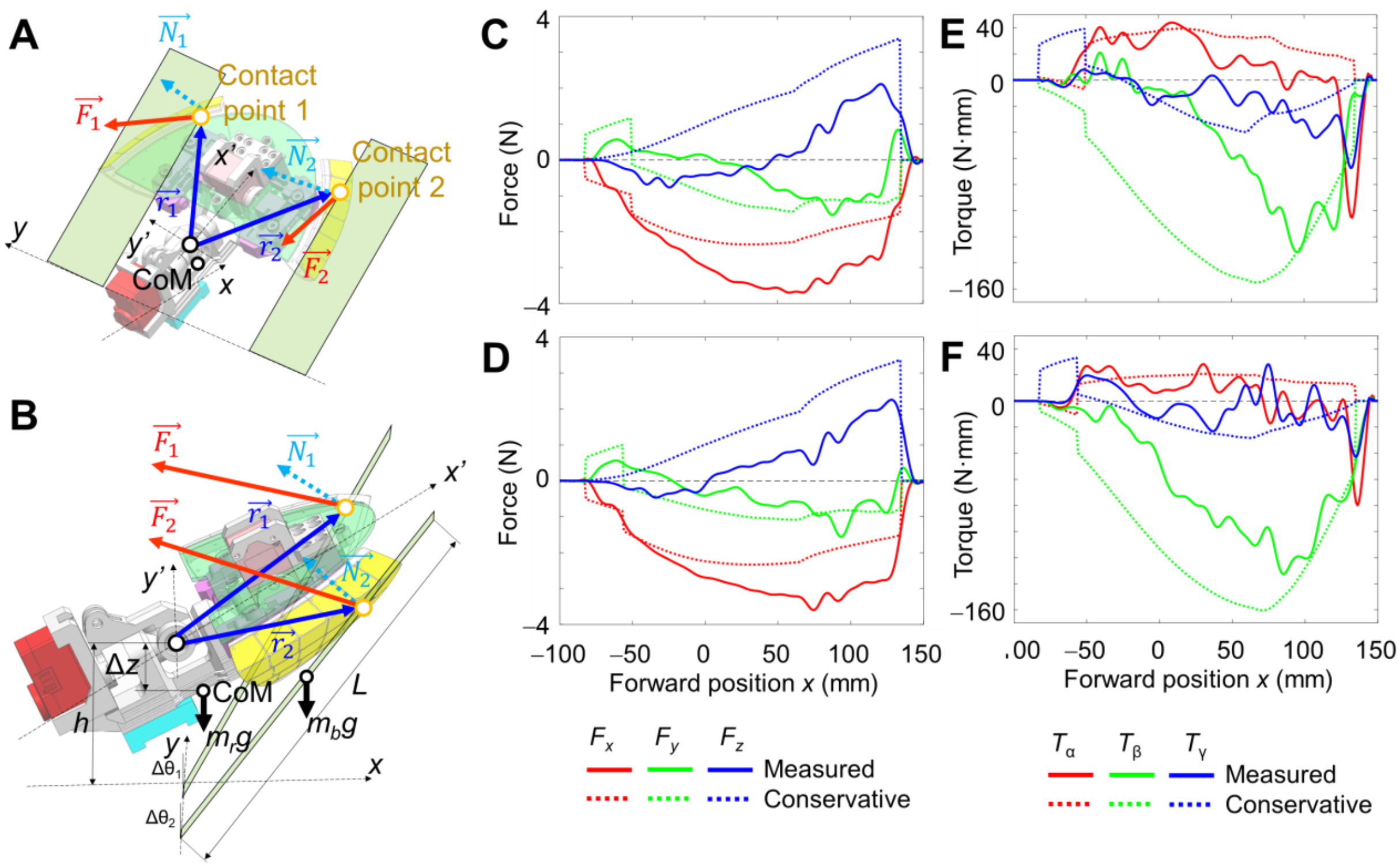

**Fig. 5. Estimating potential energy landscape gradients using measured obstacle contact forces and torques.** (**A, B**) Schematic of body-obstacle physical interaction. (A) Oblique view. (B) Side view. Red arrows show measured obstacle contact forces $\overrightarrow{F_1}$ and $\overrightarrow{F_2}$. Orange points show measured contact positions. Blue arrows show the position vector $\overrightarrow{r_1}$ and $\overrightarrow{r_2}$. Cyan dashed arrows show normal forces $\overrightarrow{N_1}$ and $\overrightarrow{N_2}$, which are measured obstacle contact forces from both beams projected into the surface normal direction of each contact point measured by the contact sensors on the face (not edge) of the body (used when subtracting friction in **Sec. S.7**). Torques resulting from measured or normal forces were cross-products of the position vectors and measured or normal forces ($\overrightarrow{T_\iota} = \overrightarrow{r_\iota} \times \overrightarrow{F_\iota}, \overrightarrow{T_{\iota,N}} = \overrightarrow{r_\iota} \times \overrightarrow{N_\iota}$), respectively. (**C–F**) Measured (C, D) forces and (E, F) resulting torques compared to conservative ones (i.e., from landscape gradients) as functions of fore-aft position ($x$) from two representative trials. (C, E) body (roll, pitch) = (0°, −10°). (D, F) body (roll, pitch) = (15°, –20°).

The conservative forces and torques (obtained as negative landscape gradients after subtracting the contribution from gravity) should be approximately equal to the sum of external ones, including the contact ones and non-contact ones contributed from gravity (see **Sec. S.3** for derivation). To test this, we compared the measured contact forces and torques with the conservative forces and torques along the relevant degrees of freedom (i.e., fore-aft translation, roll, and pitch) calculated from first principle (**Sec. 4.4**) (**Fig. 5C-F**, measured vs. conservative). We found reasonably small relative errors of $\varepsilon_x$ = 19% ± 10% (mean ± standard deviation, s.d.) in the fore-aft ($x$) direction, $\varepsilon_\alpha$ = 27% ± 57% in the roll ($\alpha$) direction; and $\varepsilon_\beta$ = 31% ± 20% in the pitch ($\beta$) direction (see **Sec. S.9** for definitions) (also see **Video 3** for relative errors across the range of state space tested). These results showed that, despite not being fully conservative, the measured obstacle contact forces and torques enabled estimation of the landscape gradients.

## 2.4 Estimated potential energy landscape matched first principle

To test whether obstacle contact force and torque sensing can be used to estimate the potential energy landscape, we estimated the landscape (see **Sec. 4.5**, **Sec. S.4** for the algorithm) over a wide range of state space (fore-aft position $-100$ mm $\leqslant x \leqslant$ 100 mm, roll 0° $\leqslant \alpha \leqslant$ 40°, pitch –40° $\leqslant \beta \leqslant$ –10°, see **Sec. S.9** for range selection) (**Fig. 6A**, **Video 2**). This range of roll and pitch was chosen as it encloses the saddle point of the barrier between the pitch and roll basins throughout the entire range of $x$ tested.

The potential energy landscape was approximated as a scalar potential function whose negative gradients best matched the measured obstacle contact forces and torques, after considering the effects of gravity. To examine how the estimated landscape matched that calculated from first principle (**Sec. 4.4**) (**Fig. 6B**), we compared the two landscapes and their gradients. We found particularly small relative errors of $\varepsilon_{PE}$ = 9% ± 6% in potential energy and reasonably small relative errors of $\varepsilon_{Grad}$ = 21% ± 13% in gradients (see **Sec. S.9** for definitions) (also see **Video 3** for relative errors across the range of state space tested). As a comparison, we also estimated the potential energy landscape using vision-based geometry sensing by assuming that beams are rigid (see **Sec. S.5** for calculation), which had a much poorer estimation accuracy (relative error $\varepsilon_{PE}$ = 106% ± 57% in potential energy and $\varepsilon_{Grad}$ = 555% ± 16155% in gradients). These results demonstrated that sensing obstacle contact forces and torques enabled the estimation of the potential energy landscape, even though they are not exactly conservative forces.

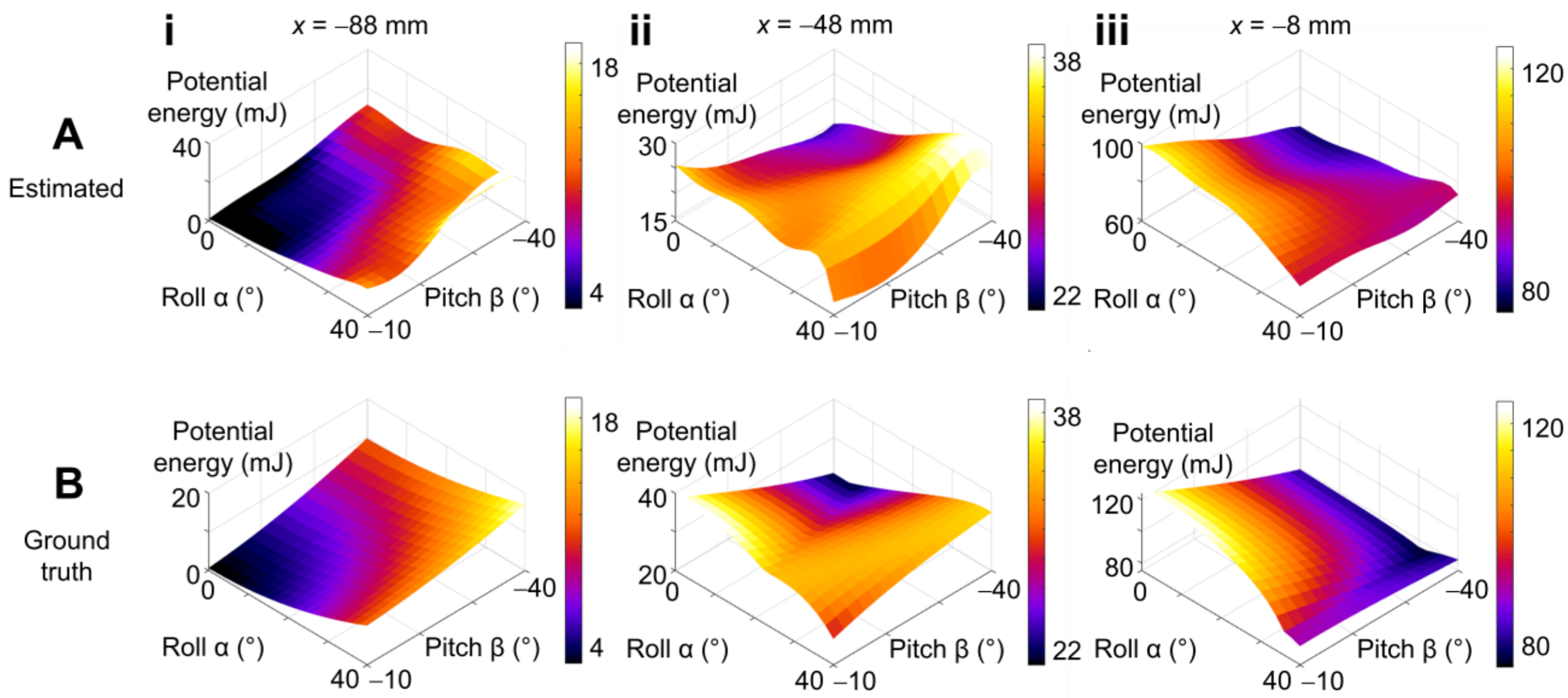

**Fig. 6. Potential energy landscape estimation.** (**A**) Estimated potential energy landscape and (**B**) that calculated from first principle at a few representative fore-aft positions: (**i**) $x$ = –88 mm, (**ii**) $x$ = –48 mm, and (**iii**) $x$ = –8 mm.

### 2.5 Bio-inspired oscillatory sensor motions improved estimation of landscape gradients

To test whether bio-inspired oscillatory sensor motions improved the estimation of landscape gradients, we conducted the same experiments in which the robot repeatedly flexed its head. We speculated that the improvement would occur only when the contact was on the edge of the robot surface (edge contact case, **Fig. 4B**, orange) rather than the faces (face contact case, **Fig. 4B**, yellow), because the friction is opposite to the relative velocity. During edge contact, the back-and-forth oscillatory motion generated a pair of relative velocities that were in opposite directions, leading to the friction canceling itself out. In contrast, during face contact, oscillations primarily resulted in tapping against the obstacle, so friction did not cancel itself out.

To test this, we compared the measured contact forces and torques (after applying a lowpass filter to reduce the force and torque oscillation from self-motion, **Sec. S.9**) with the conservative forces and torques along relevant degrees of freedom of the potential energy landscape (i.e., fore-aft translation, roll, and pitch) after subtracting contribution from gravity, in both edge and face contact cases (**Fig. 7A−F**). The robot could identify the contact cases from the contact sensors. Both contact cases occurred frequently (face: 56% ± 16%, edge: 44% ± 16% of the time among all trials).

In the edge contact case, the measured fore-aft force and pitch torque with oscillatory head flexion applied better matched the conservative forces and torques than that without (**Fig. 7G, I**), with significantly and substantially smaller relative errors from $\varepsilon_x$ = 10% ± 8% to 6% ± 4% in the fore-aft ($x$) direction ($P$ < 0.001, Student's $t$-test, Cohen's $d$ = 0.62; see **Sec. S.9** for statistical tests); and from $\varepsilon_\beta$ = 43% ± 18% to 22% ± 10% in the pitch (β) direction ($P$ < 0.001, Student's $t$-test, Cohen's $d$ = 1.44). On the other hand, in the face contact case, we found no such improvements, with significantly but not substantially different in the fore-aft ($x$) direction ($\varepsilon_x$ = 23% ± 7% vs. 22% ± 10%, $P$ < 0.001, Student's $t$-test, Cohen's $d$ = 0.21) and in the pitch (β) direction (from $\varepsilon_\beta$ = 20% ± 14% to 22% ± 18%, $P$ < 0.001, Student's $t$-test, Cohen's $d$ = –0.17). In the roll (α) direction, we found significant but not substantial improvements in the edge contact case ($\varepsilon_\alpha$ = 11% ± 11% vs. 12% ± 10%, $P$ < 0.001, Student's $t$-test, Cohen's $d$ = –0.11) and no significant improvements in the face contact case (30% ± 70% vs. 30% ± 19%, $P$ = 0.98, Student's $t$-test) (**Fig. 7H**).

These results supported our speculation that bio-inspired oscillatory sensor motions improved the estimation of landscape gradients. Besides using bio-inspired oscillatory sensor motions, we also assessed whether the estimation of landscape gradients could be improved by estimating the contact friction using the normal direction of contact (**Sec. S.7**).

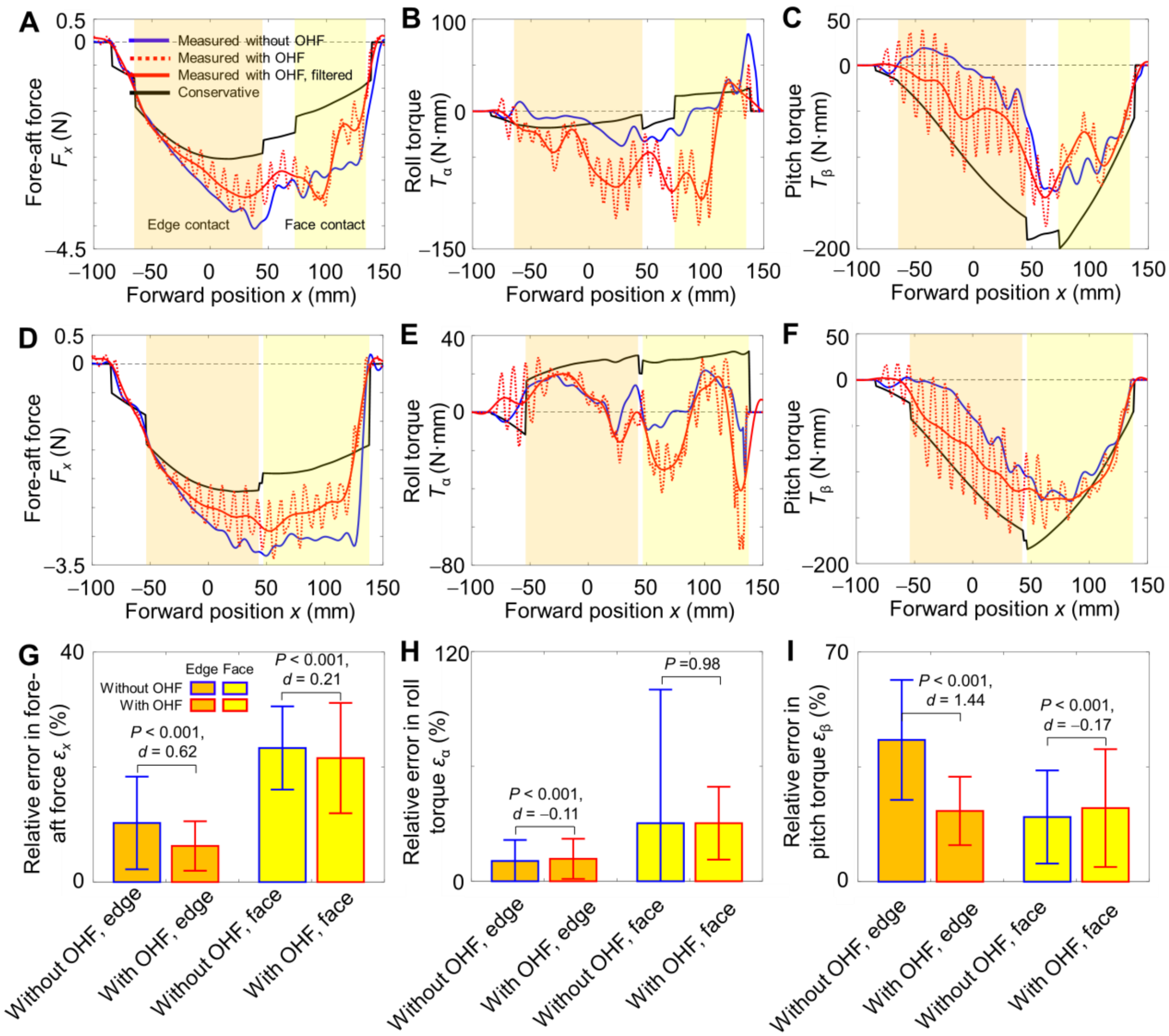

**Fig. 7. Bio-inspired oscillatory sensor motions improved estimation of landscape gradients.** (**A–F**) Measured (A, D) fore-aft force $F_x$, (B, E) roll torque $T_\alpha$, and (C, F) pitch torque $T_\beta$ compared to conservative ones (i.e., from landscape gradients) as functions of fore-aft position $x$ from a representative trial. (A–C) body (roll, pitch) = (0°, –10°). (D–F) body (roll, pitch) = (15°, –20°). Blue, red dot, red solid, and black curves show measurements without oscillatory head flexion (noted as OHF in the figure), with oscillatory head flexion, with oscillatory head flexion and filtered, and conservative forces and torques, respectively. Orange and yellow shadow shows edge and face contact cases, respectively. (**G–I**) Relative error ε of measured force and torques. Blue and red edge cases show data with or without head oscillation,

respectively. Orange and yellow show data from the edge and face contact cases, respectively. Bars and error bars are means ± 1 standard deviation (s.d.) of data across the range of state space of (fore-aft position, roll, pitch) tested. *P* is the p-value from Student's *t*-test. *d* is the Cohen's *d*.

## 3 DISCUSSION

### 3.1 Summary of contributions

We developed a new method to sense environmental physical interactions and estimate the potential energy landscape over a wide range of parameter space, providing a means to model and understand how to utilize and control physical interactions to facilitate locomotor transitions and traverse densely cluttered obstacles. We demonstrated the effectiveness of our method in a model system involving the traversal of grass-like beam obstacles. We created a robot capable of accurately sensing obstacle contact forces and torques over a wide range of body rotations experienced during traversal. Surprisingly, sensing these non-conservative contact forces and torques enabled the accurate estimation of the potential energy landscape from its gradients, even though the landscape gradients only account for conservative forces and torques during locomotor–obstacle interaction. (Indeed, the non-conservative forces not captured by the potential energy landscape gradients likely accounted for much of the relative errors observed, especially over some parts of the parameter space tested, **Sec. 2.3**, **Sec. 2.4**, **Sec. 2.5**, **Fig. 5C-F**, **Fig. 6**, **Fig. 7**, **Video 3**. Nevertheless, the estimated potential energy landscape overall was surprisingly accurate.) Cancelling friction by applying bio-inspired oscillatory sensor motions further improved estimation accuracy. Our results obtained in a minimalistic robot designed for lab experiments with systematic parameter variation provided a foundation for further refining these methods for use in free-locomoting robots (**Sec. 3.2**). Our work is a step towards establishing the potential energy landscape approach for robots to sense and use environmental physical interaction for obstacle traversal, analogous to the artificial potential field approach for robots to sense and use environmental geometry for obstacle avoidance.

Our custom force sensors and distributed contact sensors have several advantages over other sensors. Compared with commercial sensors, these custom sensors were low-cost (~$200 for all sensors and the DAQ board, compared with ~$8,000 for a commercial transducer) and easy to repair (~10 minutes to repair one worn contact sensor). Also, our distributed contact sensor design was robust against damage. Damage to a single contact sensor did not affect force sensing and only caused a small error in contact point sensory information, which could be easily fixed through interpolation. Compared with flexible sensory arrays that only provide normal forces and have high latency and nonlinearity (Luo *et al.*, 2017; Ha, Lim and Ko, 2018; Lee *et al.*, 2019; Yao *et al.*, 2020; Zhu *et al.*, 2020; Cholleti *et al.*, 2021; Roberts, Zadan and Majidi, 2021; Ramesh, Fu and Li, 2022), our force sensors measured obstacle contact forces in 3D with high linearity (**Fig. 4F-H**), and our contact sensors can operate at a high frequency due to their digital output.

Our method of sensing obstacle contact forces and torques and using them to estimate the potential energy landscape is suited for unknown environments. Specifically, we proposed that a robot estimate the landscapes from sensed contact forces and torques, which are ubiquitous in environmental interaction regardless of obstacle types and density. The 3-D forces and resulting torques covered the full degree of freedom of a rigid, single-segmented robot, which could enable landscape estimation in any desired dimensions. Our method was also verified in another robot with a different body shape (**Sec. S.8**).

Our bio-inspired oscillatory sensor motions improved the estimation of the landscape gradients when the robot body contacted the beam obstacles at the sharp edge, because it allowed friction from obstacle contact forces to cancel in the time average. This highlights another beneficial aspect of active oscillation aside from facilitating the overcoming of energy barriers (Othayoth, Thoms and Li, 2020; Othayoth *et al.*, 2021) and challenges the conventional view that such oscillations are merely wasteful motions. This also inspires a deeper understanding of the biological behavior of active oscillatory head flexion while traversing clustered obstacles (Wang, Othayoth and Li, 2022).

### 3.2 Applicability to free locomotion in arbitrary cluttered terrain

Given the advancements demonstrated here on a minimalistic, externally driven robot, challenges remain in applying our method of estimating potential energy landscapes to robots during free locomotion in arbitrary cluttered terrain.

First, our contact sensor developed for systematic lab experiments required the obstacle surface to be conductive (**Sec. 4.1**), which cannot be used in real terrain. This can be addressed by replacing our contact sensors with recently developed distributed tactile sensors (Yang *et al.*, 2019; Lei *et al.*, 2024; Liu *et al.*, 2024; Xu *et al.*, 2024) and leveraging new sampling strategies (Slepyan *et al.*, 2025) that accelerate distributed contact sensing. Future studies should adapt these sensors to small-legged robots and improve their robustness against collision and fouling.

Second, the measured forces can be substantially noisy due to non-conservative components, e.g., inertial forces, oscillations from cyclic leg propulsion, and impulses from frequent collisions. Because traversing cluttered, large obstacles is highly strenuous, a free-running robot needs to continually push against and intermittently collide with them (Othayoth *et al.*, 2021), making it difficult to build up high momentum. As a result, it usually operates in a low-speed, small-acceleration regime, where damping and inertial forces are likely small. Therefore, it is likely that conservative forces and friction dominate the noisy obstacle contact forces during free locomotion, similar to our idealized lab experiments. Future studies should test self-propelled legged robots (rather than externally driven) to produce natural locomotor-environmental physical interaction and examine the dominance of conservative forces.

Third, a robot usually does not experience the entire state space of interest; it should leverage limited sensory data along its trajectory to estimate the potential energy landscape and traverse. Future studies may employ bio-inspired strategies to identify local landscapes and determine the direction for transitioning to another basin using limited data. For example, cockroaches (Bell, Roth and Nalepa., 2007; Li *et al.*, 2015; Othayoth, Thoms and Li, 2020; Wang, Othayoth and Li, 2022) and ground beetles (Thiele, 1977) often wedge in between obstacles like shrubs or rock cracks to seek shelter. They may leverage these exploratory motions to sample obstacle contact forces and torques around their neighboring area (i.e., system states) to identify weaker spots (assuming that they have such a preference). This process could be simplified and encoded in the animal's neural network (Mongeau *et al.*, 2015; Cellini *et al.*, 2024). Similarly, a robot can actively sample the neighboring state space around its current state (53) to estimate a local potential energy landscape and then identify and follow the least-resistant trajectory (54, 55) to make transitions across landscape basins, using the right locomotor modes to traverse obstacles.

### 3.3 Implications for biology

The oscillatory sensor motion improved the estimation of landscape gradients, which may benefit a robot when identifying the direction of transition with limited sensing capabilities. The usefulness of this bio-inspired motion, in return, advanced our understanding of the biological active motion. Previous biological studies on contact-based active sensing behavior have shown that animals' force sensing inherently involves sensor motion, i.e., animals often move their sensors to enhance sensing (Prescott, Diamond and Wing, 2011; Bajcsy, Aloimonos and Tsotsos, 2018). For example, when encountering an obstacle, an insect uses its antennas to repeatedly touch the obstacle (e.g., cockroach (Okada and Toh, 2006; Baba, Tsukada and Comer, 2010)) or uses an antenna search-and-sample behavior aided by body and head rotating (e.g., stick insect (Schütz and Dürr, 2011; Berendes and Dürr, 2022)), which helps locate the obstacle and induces turning to avoid collision (Baba, Tsukada and Comer, 2010; Berendes and Dürr, 2022; Othayoth *et al.*, 2022). Similarly, a rat actively moves its whiskers back and forth against objects when exploring the environment (Hartmann, 2001, 2011; Bush, Solla and Hartmann, 2016; Zweifel and Hartmann, 2020). Inferring from comparative robotic studies (Schultz *et al.*, 2005; Solomon and Hartmann, 2010; Emnett, Graff and Hartmann, 2018), such active sensing behavior probably enables the animal to extract object contours (Huet, Rudnicki and Hartmann, 2017; Huet, Emnett and Hartmann, 2022). Similarly, the discoid cockroach's oscillatory head flexion in beam obstacle traversal (Othayoth, Thoms and Li, 2020; Wang, Othayoth and Li, 2022) may be a novel form of active sensing and helps the animal (or a robot) estimate the local trend (e.g., less resistive along specific directions) of environmental physical interaction encoded in the potential energy landscape and identify the direction of least-resistance transition towards a more favorable landscape basin (i.e., locomotor mode).

Besides active head flexion, when traversing cluttered large obstacles, the cockroach's body had obvious oscillation in all translational and rotational degrees of freedom, due to legged self-propulsion (Li *et al.*, 2015; Othayoth, Thoms and Li, 2020; Wang, Othayoth and Li, 2022). Previous studies have shown that body oscillation provides kinetic energy fluctuations to the system and facilitates overcoming the energy barrier to transition to a desired locomotor mode (Othayoth, Thoms and Li, 2020; Othayoth *et al.*, 2021). Our study here further suggested that these body oscillations can help exploratorily sample the state space around the current state. Further studies should examine this by applying body oscillations to observe how they modulate the force and torque sensing.

### 3.4 Capturing stochastic dynamics with potential energy landscapes

Our experimental result showed that, for a robot traversing cluttered large obstacles, the conservative forces and torques (from landscape gradients) dominated the interaction, while non-conservative forces and torques, such as those from friction, still contributed. Future studies should model the dynamics of such self-propelled locomotor–obstacle interactions as potential energy landscape-dominated, stochastically perturbed dynamics with diffusion (i.e., Langevin dynamics (66–68)), where friction, damping, inertia effects, and kinetic energy fluctuations (Othayoth *et al.*, 2021) are also considered. This will enable model prediction of system dynamical trajectories, which can not only allow model-based control of robot traversal but also improve our understanding of biological motion.

## 4 MATERIALS AND METHODS

### 4.1 Sensor design and calibration

Force sensors. Each custom 3-D force sensor consisted of three load cells (BF-02088B, HK Bingf Sci. & Technol. Corp., China) serially connected and orthogonal to one another to measure the forces along the body $x$'-, $y$'-, and $z$'-axes (**Fig. 4B**, magenta). Each load cell had a range of ± 20 N and a precision of ± 0.004 N, which suited our experiment where the obstacle contact forces were < 10 N. The obstacle contact forces were read from the load cells by amplifier chips (HX711, Avia Semiconductor, China) on the DAQ board (**Fig. 4B**, green).

To calibrate each force sensor, we positioned it with the $\pm x$, $\pm y$, or $\pm z$ direction facing downward and secured the end that was to be attached to the inner core of the robot's body. For each of these six sensor orientations, we zeroed the sensor readout, then applied a force by hanging a weight of 1 to 5 N with an increment of 1 N and recorded the readouts in all three relevant degrees of freedom. To quantify the precision and crosstalk, we performed linear regression (**Fig. 4F**) of the readouts on the applied forces on the same or different relevant degrees of freedom, respectively.

Contact sensors. The contact sensors on the faces were made of copper tape (Qian and Goldman, 2015), and those on the edges were made of 0.5 mm diameter aluminum wires (**Fig. 4B**, yellow and orange, **Fig. 4D, E**). These sensors were connected to capacitive touch detection chips (MPR121, Freescale Semiconductor, TX) on the DAQ board (**Fig. 4B**, green). Because the beam surface was grounded, the capacitive touch detection chips detected a voltage drop (i.e., a digital falling edge signal) when a sensor touched the beam surface.

To achieve high accuracy in torque measurements, we must accurately measure contact position on the robot body. We estimated the contact position as the center of each contact sensor that detected contact forces. In this case, the worst estimation should occur when the contact occurred at the corners of sensors on the face or endpoints of sensors on the edge, because these points are the farthest from their respective centers. Thus, we defined the maximal position error of each contact sensor as the maximal distance from the corners or endpoints to the center. We then carefully designed the pattern of these contact sensors to minimize this error. Because the left and right shells are symmetric, we first designed the sensor pattern on the right shell and mirrored it to the left. The number of sensors should be smaller than 36 on each piece of shell, limited by the port number from the DAQ board. To minimize position error for a given sensing area, each sensor on the face was designed to be "squared" (i.e., each had sides of roughly the same length), roughly the same size, and placed contiguously without interstitial spaces. After multiple iterations of designing and testing, we arranged the sensors into a grid of 6 (tangential) × 3 (radial) on the bottom face, 6 × 1 along the edge, and 6 × 1 on the top face (**Fig. 4I**).

### 4.2 Experimental protocol

*Experiment without oscillatory head flexion.* Before each trial, the robot body was positioned at a distance of 200 mm ($x$ = –200 mm) from the beams, at a height of $z$ = 138 mm from the beams' bottom edges, in the middle of the two beams (lateral position $y$ = –6 mm), and pointing forward (body yaw $\gamma$ = 0°). The robot was rotated to the desired roll and pitch and kept at that orientation throughout each trial. The beams were set vertically with a 130 mm wide gap symmetric to the midline (so that the robot body needed to roll ~40° to avoid touching them). All the force sensors were zeroed, and the LabVIEW program started data recording. Then the robot was moved forward at a constant speed of 20 mm·s$^{-1}$ by a distance of 500 mm, which guaranteed the robot fully passed the beam obstacles, after which the beams bounced back to vertical. Finally, we stopped data recording and moved the body back to its initial position.

To estimate a small region of the potential energy landscape, we varied the robot body roll and pitch within it. Specifically, we varied the roll $\alpha$ from 0° to 40° in increments of 5° and the pitch $\beta$ from –10° to –40° in decrements of 5° (note that negative pitch meant pitching the body upward). At each combination of roll $\alpha$ and pitch $\beta$, we performed five trials, resulting in a total of $n$ = 315 trials. Note that because the robot's frames and links were not stiff, the body roll and pitch changed slightly (maximum roll change < 10°, maximum pitch change < 5°) in each trial. We measured the roll and pitch from the IMU to account for this.

*Experiment with oscillatory head flexion.* In the experiment to test whether oscillatory head flexion allowed friction to cancel, we followed the same protocol as above, except that we further commanded the robot to flex its head repeatedly after the LabVIEW program started recording data. The head was flexed with an amplitude of 20°, centered with the head pointing along the long axis of the body at a frequency of 2 Hz (see **Sec. S.6** for frequency selection). To randomize the oscillatory head flexion phase to test if the friction cancellation is sensitive to phase, in each trial, the robot started moving forward after a random period of up to one flexion period (following a uniform distribution).

### 4.3 Force analyses

For either beam, the obstacle contact force (**Fig. 5A, B**, red) was directly measured from the force sensors. The contact position (**Fig. 5A, B**, orange) was the center of the touch sensor. The position vector (**Fig. 5A, B**, blue) was the distance from the body's geometric center to the center. The resulting torques were the cross-products of the position vector and the measured forces. The resulting torques along the roll, pitch, and yaw directions were the projections of the 3-D torques along the roll, pitch, and yaw degrees of freedom, respectively. The total measured forces and resulting torques were the sum of those from the two beams.

### 4.4 Potential energy landscape calculation from first principle

The system's potential energy $PE$ is the sum of the gravitational potential energy of the robot body and the beams $PE_G$ and the elastic potential energy from the beams $PE_E$:

$$PE = PE_G + PE_E$$

$$PE_G = m_r g(\Delta z - h) + \frac{1}{2} m_b g L(\cos\Delta\theta_1 + \cos\Delta\theta_2),$$

$$PE_E = \frac{1}{2} k_1 \Delta\theta_1^2 + \tau_1 \Delta\theta_1 + \frac{1}{2} k_2 \Delta\theta_2^2 + \tau_2 \Delta\theta_2,$$

where $m_r = 0.53$ kg and $m_b = 0.04$ kg is the mass of the robot body and each beam, respectively, $g$ was the gravitational acceleration, $\Delta z$ is the vertical distance between the geometric center and center of mass of the robot body, $h = 8$ mm is the height of the geometric center from beams' lower edges, $L = 200$ mm is the beam length, $\Delta\theta_{1,2}$, $k_{1,2}$, and $\tau_{1,2}$ are the deflection angles, the torsional stiffness, and the preload (**Sec. S.1**) of the left and right beams, respectively. See **Fig. 5B** for definitions of parameters. Note that we have newly included the preload compared to the previous study (8), as it makes a significant contribution to the contact forces.

The potential energy landscape was the system's potential energy as a function of fore-aft position, roll, and pitch ($x$, α, β). To compare with the measured forces and torques (**Sec. 2.3, Sec. 2.5**) and the estimated landscape (**Sec. 2.4**), we also calculated the potential energy landscape and its gradients from first principle as a function of fore-aft position, roll, pitch. To obtain the potential energy, for a given robot body position and orientation, we calculated each beam's deflection angle $\Delta\theta_{1,2}$ from the geometry of the system based on the fact that the beam cannot penetrate the body. The deflection angles were set to zero before and after the robot body touched and detached from the beams. Then, the potential energy was calculated using the equations above. Note that in the case of oscillatory head flexion, we still assumed the potential energy landscape to be the same as that of a static head because a previous study demonstrated that oscillatory head flexion only affects the landscape by a small degree (Wang, Othayoth and Li, 2022).

We calculated the conservative forces $\boldsymbol{F}_{\mathbf{con}}$ by differentiating the potential energy landscape along fore-aft position ($x$), roll (α), and pitch (β), using a central difference method with a perturbation of $d = 10^{-4}$ mm or rad:

$$F_{\mathrm{con},q\mathrm{i}} = -\,[PE(q_\mathrm{i} + d) - PE(q_\mathrm{i} - d)]/2d,\ q_\mathrm{i} \in \{x, \alpha, \beta\}.$$

### 4.5 Potential energy landscape estimation

We estimated the potential energy landscape using the data from the experiments without oscillatory head flexion. The average measured fore-aft force and roll and pitch torques were treated as negative landscape gradients and processed by a meshless Helmholtz-Hodge decomposition (Patane, 2022) (see **Sec. S.4** for algorithm).

## ACKNOWLEDGMENTS

We thank Ratan Othayoth, Qihan Xuan, Shai Revzen, Noah Cowan, Jean-Michel Mongeau, Jeremy Brown, Mitra Hartmann, Ioannis Kevrekidis, and Anastasia Georgiou for discussion and Xiao Yu for help in assembling the experimental setup and preliminary testing.

## AUTHOR CONTRIBUTIONS

Yaqing Wang: Conceptualization, Formal analysis, Investigation, Methodology, Software, Visualization, Writing – original draft. Ling Xu: Investigation, Methodology. Chen Li: Conceptualization, Funding acquisition, Supervision, Writing – review & editing.

## STATEMENTS AND DECLARATIONS

**Ethical considerations:** Not applicable.

**Consent to participate:** This article does not contain any studies with human or animal participants. There are no human participants in this article, and informed consent is not required.

**Consent for publication:** Not applicable.

**Declaration of conflicting interest:** The authors declared no potential conflicts of interest with respect to the research, authorship, and/or publication of this article.

**Funding statement:** This work was supported by a Beckman Young Investigator Award from the Arnold and Mabel Beckman Foundation for CL, a Career Award at the Scientific Interface from Burroughs Wellcome Fund for CL, a Bridge Grant from Johns Hopkins University Whiting School of Engineering for CL, and a JHU LCSR Research Experience for Undergraduates in Computational Sensing and Medical Robotics (CSMR REU) from National Science Foundation to LX.

**Data availability:** The CAD models, control codes, and experiment data are available from GitHub after publication: https://github.com/TerradynamicsLab/landscape_reconstruct.

**SUPPLEMENTAL MATERIAL**

## S.1 Robotic system design and data acquisition

The experiment system consisted of an actuated robot body and two flexible beams (**Fig. 4A**, **Fig. S1A**). To best compare with the previous study (Othayoth, Thoms and Li, 2020), the principal axis lengths (156 mm length × 168 mm width × 62 mm height) of the uncropped full shape of the body (**Fig. 4B**, transparent green) were maintained the same as those in the previous design (Othayoth, Thoms and Li, 2020). The body rotation was measured by an inertial measurement unit (IMU, BNO055, Adafruit Industries, NY) (**Fig. 4B**, cyan). Preliminary tests (**Fig. S1B)** demonstrated that the IMU provided accurate rotation measurements with errors < 8° in the roll direction and < 2° in pitch, respectively.

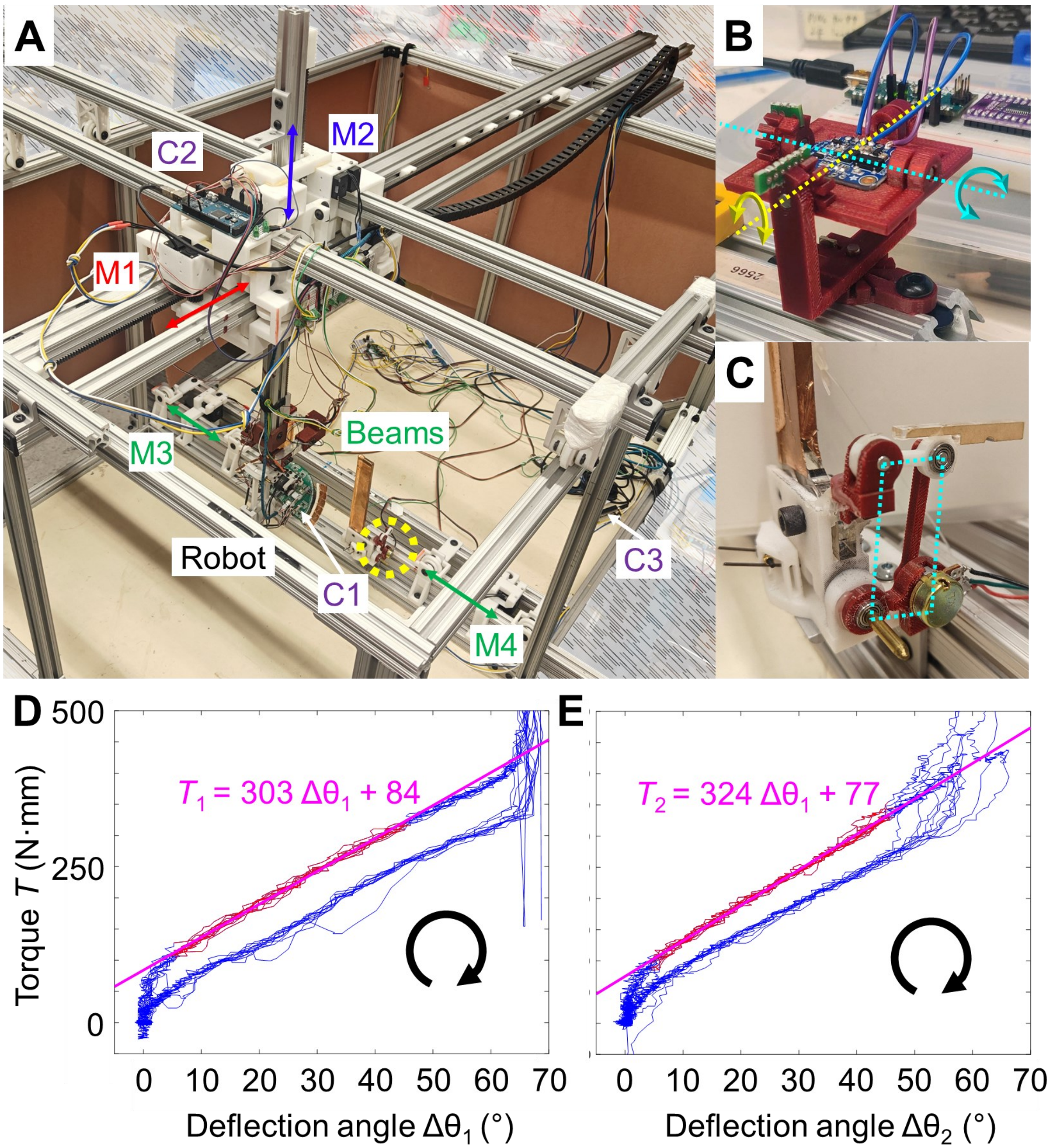

**Fig. S1. Robotic system actuation, IMU accuracy testing, and beam calibration.** (**A**) Robotic system actuation structure from an oblique top view. Purple shows microcontrollers (C1: Teensy 4.0, C2: Arduino Mega, C3: OpenCM 9.04). Red, blue, and green show servo motors. (**B**) IMU accuracy testing tool to systematically vary roll (yellow) and pitch (cyan). (**C**) Potentiometer and parallel four-bar linkage (cyan) to measure beam angles (dashed yellow circle in A). (**D, E**) Beam torque as a function of beam deflection

angle from (D) left and (E) right beams. Arrows show curve directions during multiple loading (higher torque) and unloading (lower torque) tests. Red portion of the data during loading was used for beam stiffness and preload calculations. Magenta lines and equations show linear regression results.

The two beams were made using the same method as in (Othayoth, Thoms and Li, 2020) (**Fig. 4A**, green). Each beam was a rigid acrylic plate (200 mm height × 30 mm width × 6 mm thick) attached to the base via a 3-D printed torsion spring joint. We only allowed the beams to deflect forward (positive deflection angles $\Delta\theta_{1,2}$ in **Fig. 5B**) by pulling them with a fishline attached to their backs, thereby preventing backward deflection (negative deflection angles $\Delta\theta_{1,2}$). To measure the beam deflection angles $\Delta\theta_{1,2}$, we attached a potentiometer (100 K Ohm Potentiometer, HiLetgo, China) to each beam's rotational joint via a parallel four-bar linkage (**Fig. S1C**). The two end terminals were powered at 5 volts, and the voltage at the wiper was measured to calculate the rotation angle with a resolution of 0.3°.

We chose a high beam stiffness of ~300 $N \cdot mm \cdot rad^{-1}$ (Othayoth, Thoms and Li, 2020) so that the beam contact forces were maximal to maximize the force sensors' signal-to-noise ratio and the pitch-to-roll transition barriers on the potential energy landscape were obvious. We characterized the torsional stiffness and preload of the beams by measuring the restoring torque as a function of the deflection angle using a force sensor (Optoforce OMD-20-FG, OnRobot, Denmark, range: 40 N compression only, resolution: 0.002 N), similar to (Othayoth, Thoms and Li, 2020). To do this, the beam joint axis was fixed. The bottom of the joint was supported by the force sensor with a moment arm of 100 mm. We slowly pushed the beam and recorded the deflection angle and support force. The torque was calculated as the product of the moment arm and the measured force. We performed 11 trials for each beam. Either beam's torsional stiffness and preload (i.e., the minimal torque to deflect) were calculated as the slope and the intercept of the linear regression of the torque on the deflection angle (**Fig. S1D, E**). We only used the data of the deflection angles between 5° and 45° and pushing forward (**Fig. S1D, E**, red). We found that the left beam had a stiffness of 303 $N \cdot mm \cdot rad^{-1}$ and an 84 $N \cdot mm$ preload, and the right beam had a stiffness of 324 $N \cdot mm \cdot rad^{-1}$ and a 77 $N \cdot mm$ preload.

The robot body's fore-aft and vertical displacement and the beam bases' lateral displacement were actuated by servo motors (DYNAMIXEL XM430-W210-T, ROBOTIS Co., South Korea) via a custom gear-rack mechanism (**Fig. S1A**, red, blue, and green arrows, respectively), whose encoders measured the displacements with a precision of 0.01 mm. The robot body's roll, pitch, and oscillatory head flexion were actuated by servo motors (DYNAMIXEL XC330-M288-T, ROBOTIS Co., South Korea) (**Fig. 4B**, red, blue, and pink, respectively), whose encoders measured the rotation angles with a precision of 0.1°.

The body rotation, contact forces, and contact positions (**Sec. 4.3**) were collected by the DAQ board (**Fig. 4B**, green). The large number of sensors (6 force sensors and 60 contact sensors) imposed a heavy demand on onboard data acquisition while being constrained by the limited space in the robot body. Our early attempts using commercial breakout boards of the data acquisition chips (i.e., HX711 and MPR121) presented severe space arrangement and wiring challenges (see **Sec. S.2** for iterative revision), leading us to integrate them into a custom DAQ board. We designed a DAQ board (**Fig. 4B**, green) with a microcontroller (Teensy 4.0, PJRC, OR), acquisition chips, and other common electrical components (i.e., resistors, capacitors). It was mounted in the robot's head. It allowed readouts from six load cells, 60 contact sensors, and an inertial measurement unit (IMU) and relayed them to a LabVIEW interface at 50 Hz. Each load cell was connected to the DAQ board via a bundle of four wires. Each contact sensor was connected via a single thin (30 AWG) jumping wire and a pull-up resistor. The IMU was connected via a bundle of four wires. The shape of the DAQ board was half an ellipse to fit in the head, with a rectangular hole in the middle to accommodate a servo motor (**Fig. 4B**, pink). To avoid interfering with force sensing, there was a minimum distance of 2 mm between any shells and force sensors, and all connection wires from the load cells were intentionally left loose. For all the electrical components, we chose surface-mount device (SMD) models and mounted them on the top of the PCB board to avoid being shorted by the aluminum load cells beneath. We designed the DAQ board using EAGLE (Autodesk, Inc., CA, USA) and manufactured it (PCB prototyped and assembled) via JLCPCB, China, which cost approximately $20 per unit when ten units were purchased together.

The beam angles were collected by a microcontroller (Arduino Mega, Arduino, Italy). All the servo motors were commanded and reported their rotation angles to a microcontroller (OpenCM 9.04, ROBOTIS Co., South Korea). Both microcontrollers acquired data at a frequency of 50 Hz (the same as the DAQ board) and were mounted either on the moving frame or at the corner of the setup (**Fig. S1A**, C2, C3). We used a LabVIEW program to receive data from the microcontrollers, record it, and send the desired motor commands to their controllers. The LabVIEW graphical user interface (GUI, see **Video 1**) visualized these data (i.e., the robot's position, orientation, head flexion angle, contact force, and contact position on each piece of shell, and the beams' lateral positions and deflection angles) while allowing manual control of the system's actuation in real-time and conducting automatic pre-programmed experiments. During automatic experiments, the robot moved forward through the beams in each trial (**Sec. 4.2**), and then safely returned to its initial position between trials. This was ensured by first elevating the body above the top edge of the beams and then moving the beams outward, creating a gap that was significantly wider than the body. Then, the body moved backward to the initial fore-aft position. Finally, the body was lowered, and the beams moved inward to their initial positions. We used two cameras (Logitech C920 HD PRO, Logitech, Switzerland) synchronized by Open Broadcaster Software (OBS) to record the experiment from the side and isometric views at a frame rate of 30 Hz and a resolution of 960×720 pixels (see cropped videos in **Video 1** and **3**).

## S.2 Iterative development of robotic systems

The robotic system used for experiments reported in this paper (**Sec. 2.2**, **Sec. 4.1**, and **Sec. S.1**) was achieved through iterative development. In the first design (**Fig. S2**), the robot's body consisted of a head, thorax, and abdomen (**Fig. S2**, purple, pink, and dark yellow, respectively), mimicking the body parts of a cockroach. The body was cropped from a smooth, ellipsoidal shape with no sharp edges. The robot could sense contact forces and torques on its head and abdomen using two commercial force and torque sensors (ATI Mini 40, ATI Industrial Automation, NC) (**Fig. S2**, magenta). The thorax could move in fore-aft directions, actuated by the same mechanism as in **Sec. S.1**. Its roll was actuated by a servo motor (DYNAMIXEL XL430-M288-T, ROBOTIS Co., South Korea) (**Fig. S2**, red). Its pitch was actuated by two servo motors (DYNAMIXEL XL-320, ROBOTIS Co., South Korea) (**Fig. S2**, blue) on both left and right controlled in exactly mirrored motion. The head and abdomen could flex around the thorax, each actuated by a servo motor (DYNAMIXEL XL430-W250-T, ROBOTIS Co., South Korea) (**Fig. S2**, orange and yellow). Both oscillatory head and abdomen flexion were inspired by cockroaches' behavior (Wang, Othayoth and Li, 2022).

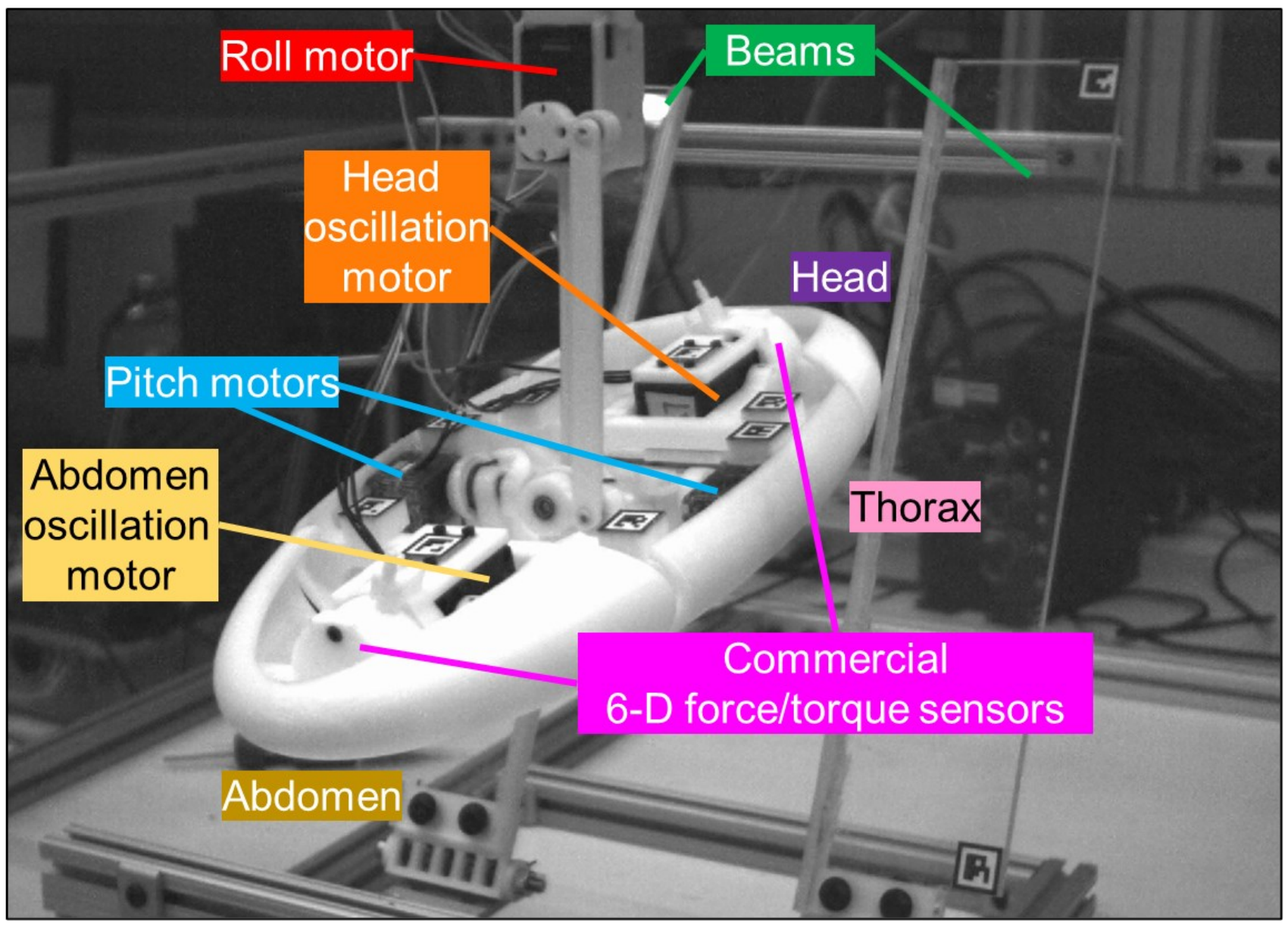

**Fig. S2. First iteration of sensor-instrumented robotic systems.** 3-D printed mockups were installed instead of commercial force/torque sensors for preliminary testing.

We found many problems with this first design. The robot body had a much larger size (due to the large ATI force/torque sensor size) and a different shape than that in the previous study (Othayoth, Thoms and Li, 2020), which made it more difficult to compare with and build on previous results. The rotation sequence of the mechanism following the previous study (Othayoth, Thoms and Li, 2020) did not match the Tait-Bryan convention (i.e., Z-X-Y vs. Z-Y-X). The pitch of the thorax was actuated by two motors that caused over-constraint. The gap between the head or abdomen and thorax frequently caught a beam and caused a large strain on the sensors. The robot could not sense the contact forces and torque on the thorax, and it could not separate the contact forces and torques with either beam. Most of these problems (except for the rotation sequence and the gap) were mainly because the commercial sensors were too bulky to be installed in a confined space. These led us to use smaller custom sensors.

The second design (**Fig. S3**) was improved from the first design. The rotation sequence of the mechanism was corrected, and the body pitch was actuated by a stronger servo motor (DYNAMIXEL XL430-W250-T, ROBOTIS Co., South Korea) (**Fig. S3**, blue). The ability to have oscillatory abdominal flexion was removed to make room for sensor chips. The body had a structure with an inner core and multiple outer shells on the head, thorax, and abdomen to embed force sensors. The thorax was separated into front and rear parts (**Fig. S3**, pink). The shells were further separated in the sagittal plane to differentiate the contact with either beam. Between each shell and the core, we added the same custom force sensor (**Fig. S3**, magenta) as in **Sec. 4.1** to measure contact forces. We installed contact sensors on the shell faces (**Fig. S3**, yellow) and covered the beam surfaces with grounded copper tape (**Fig. S3**, green) to measure contact positions. The force sensors and contact sensors were read by the same chips as in **Sec. 4.1,** but all on commercial breakout boards. The data were collected by a microcontroller (Arduino Mega, Arduino, Italy) mounted in the robot's abdomen.

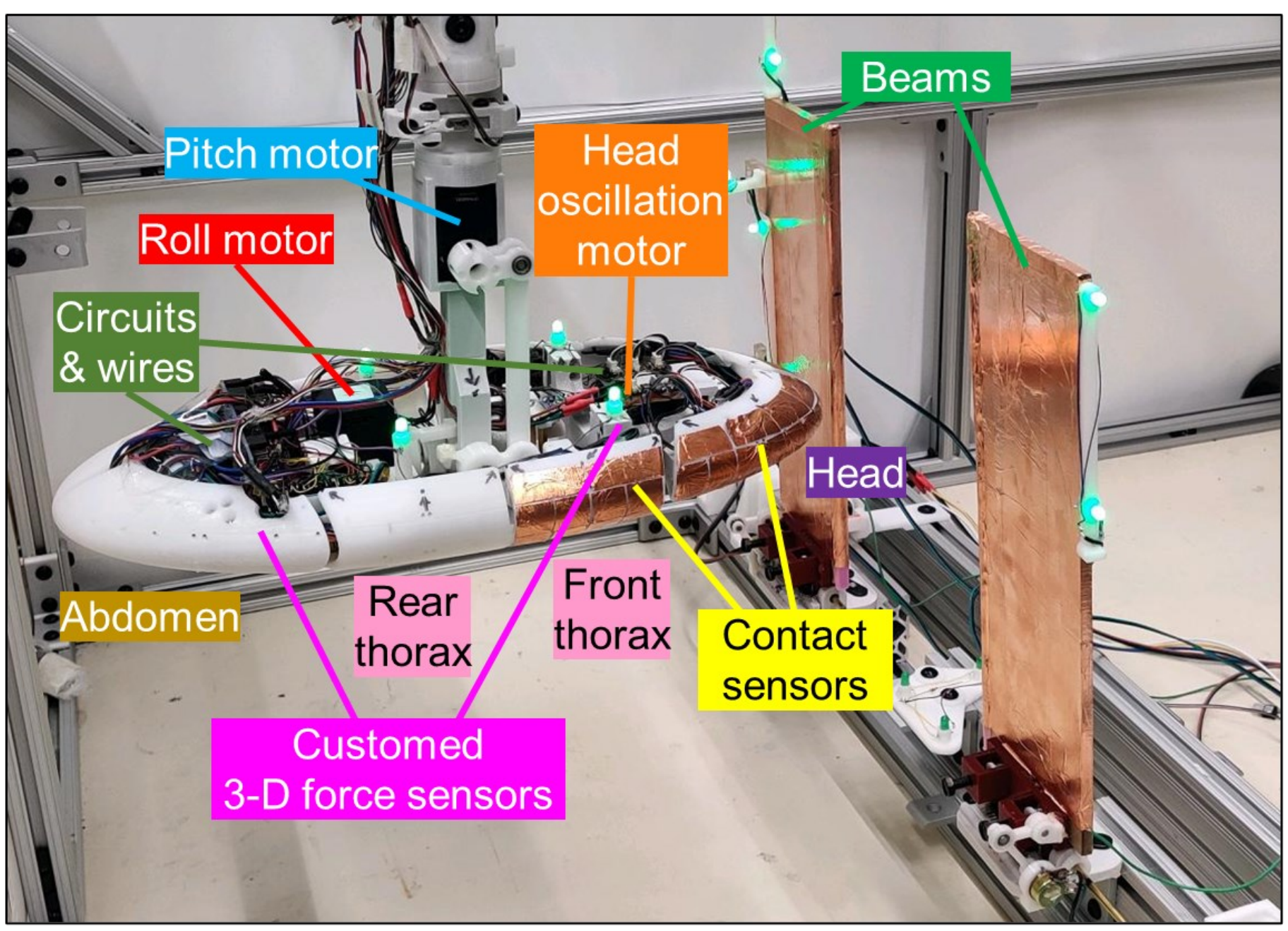

**Fig. S3. Second iteration of sensor-instrumented robotic systems.** Contact sensors were only installed on the head and front thorax for preliminary testing.

Compared to the first design, the second design enabled sensing contact forces and torques with either beam. The custom sensors were much smaller than the commercial ones, allowing us to install more of them. However, a few problems remained, such as the mismatch in size and shape with (Othayoth, Thoms and Li, 2020) and the frequent catching of the beams by gaps. Moreover, the bulky breakout boards and numerous wires (**Fig. S3**, dark green) added a large drag force to the shells and resistance to the rotation mechanism, which hindered accurate force sensing and body rotation to large angles. These led us to further design custom DAQ boards to spare more space.

The third design (**Fig. S4**) was improved from the second design. The robot body was designed to have the exact same size and shape as in the previous study (Othayoth, Thoms and Li, 2020). The front and rear thorax and abdomen were replaced by a single body (**Fig. S4**, pink). The servo motors actuating the body roll and pitch and oscillatory head flexion were replaced by a smaller model (**Fig. S4**, red, blue, and orange), the same as in **Sec. S.1**. We designed two DAQ boards that recorded the contact forces and positions on the head and body, respectively. We moved the microcontroller to the DAQ in the body to reduce wiring. We also added an IMU to record the body roll and pitch.

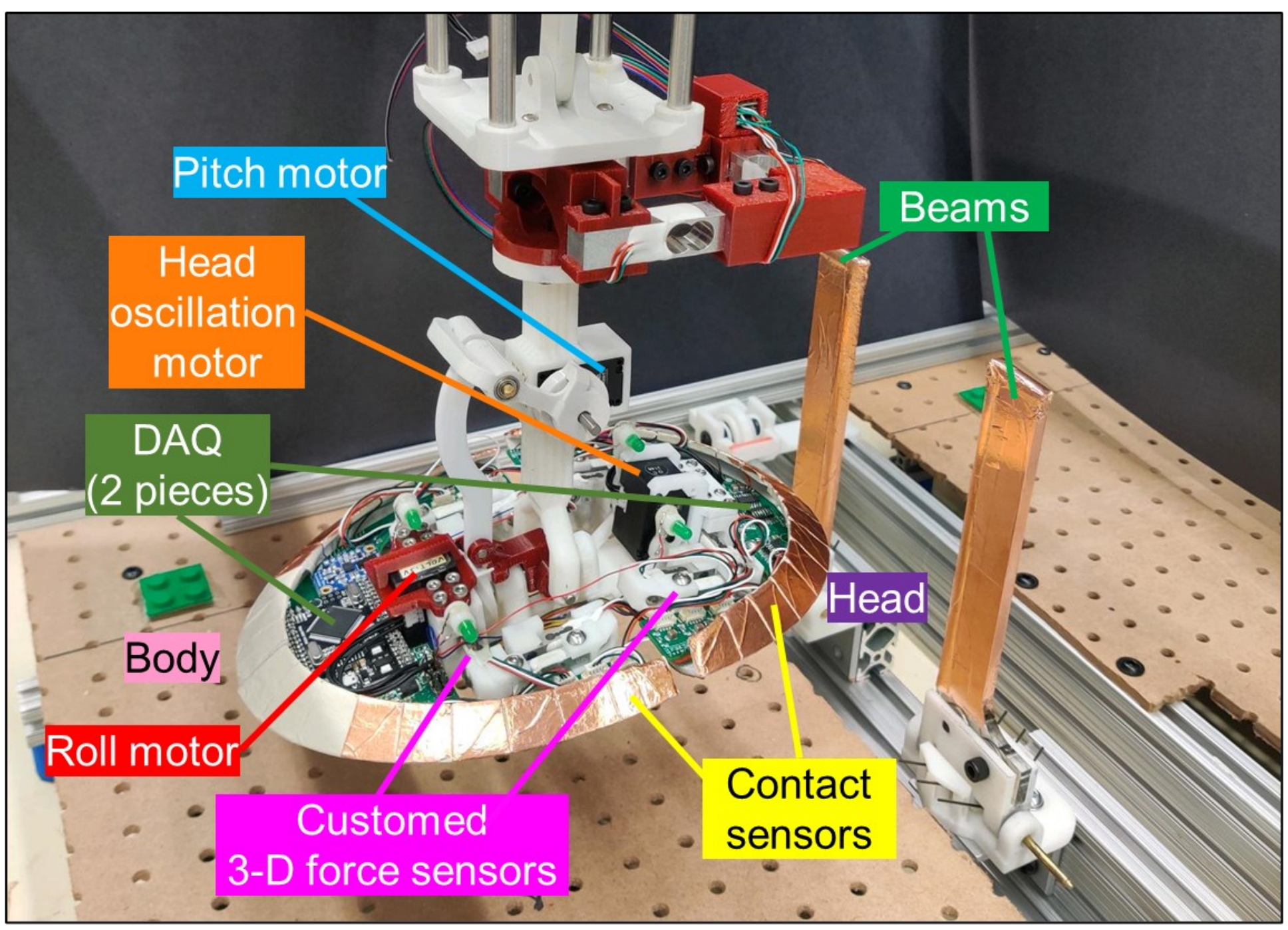

**Fig. S4. Third iteration of sensor-instrumented robotic systems.**

Compared to the second design, the third design successfully adapted the body size and shape in the previous study (Othayoth, Thoms and Li, 2020). We also minimized wiring so the robot could rotate easily to large angles. However, the problem of frequent catching of the beams by body gaps remained. Moreover, the robot could not differentiate the contact on the faces or edges, which was required to subtract friction (**Sec. S.7**). These led us to further install contact sensors on the edges in the final design. Considering that interaction between the robot and beams mostly happened when they contacted at the head, we also removed the shells of the body to avoid catching beams.

## S.3 Theoretical relation between forces and potential energy landscapes

Here, we prove that for a single rigid body interacting with obstacles, assuming quasi-static motion and frictionless contact, the obstacle contact forces and torques are negative gradients of the potential energy landscape of the system after subtracting the contribution from gravity.

The system's states can be defined as the displacements in fore-aft ($x$), lateral ($y$), and vertical ($z$) directions and rotations in roll (α), pitch (β), and yaw (γ) directions. The gradients of the system's potential energy $PE$ are the negative conservative external forces and torques $\boldsymbol{F}_{\mathrm{con}}$ along the states $\boldsymbol{q}$ = ($x$, $y$, $z$, α, β, γ):

$$\partial PE/\partial q_{\mathrm{i}} = -F_{\mathrm{con},\, q\mathrm{i}},\ q_{\mathrm{i}} \in \{x, y, z, \alpha, \beta, \gamma\}.$$

Due to the assumption of quasi-static motion, the system has no kinetic energy, and all the work $W$ done by self-propulsion (e.g., from legs) during physical interaction with the obstacles is converted into the system's potential energy:

$$W = \Delta PE.$$

The propulsive forces and resulting torques are the partial derivatives of this work:

$$F_{\mathrm{P},\, q\mathrm{i}} = \partial W/\partial q_{\mathrm{i}},\ q_{\mathrm{i}} \in \{x, y, z, \alpha, \beta, \gamma\},$$

Also, due to quasi-static motion, the external forces and torques on the robot are balanced:

$$F_{\mathrm{G},\, q\mathrm{i}} + F_{\mathrm{C},\, q\mathrm{i}} + F_{\mathrm{P},\, q\mathrm{i}} = 0,\ q_{\mathrm{i}} \in \{x, y, z, \alpha, \beta, \gamma\},$$

where $F_{G, qi}$, $F_{C, qi}$, and $F_{P, qi}$ are the gravitational, contact, and propulsive forces and resulting torques, respectively. If the robot's position and rotation are measured, the gravitational force and resulting torque (considering that the origin of the robot's body frame may not overlap with the center of mass) can be calculated. Combining all the equations above, we can calculate the contact forces and torques as the negative landscape gradients subtracting the gravitational force and resulting torque:

$$F_{C, qi} = -(\partial PE/\partial q_i + F_{G, qi}),\ q_i \in \{x, y, z, \alpha, \beta, \gamma\},$$

The assumption of quasi-static motion is nearly fulfilled in our experiment, but not in free locomotion through cluttered large obstacles (see **Sec. 3.2** for discussion). The assumption of frictionless contact is usually not fulfilled, and this led us to subtract friction to improve the estimation of landscape gradients (**Sec. 2.4**).

## S.4 Meshless Helmholtz-Hodge decomposition

We estimated a potential energy landscape as a scalar function whose gradients best matched the given vector field of estimated gradients (**Fig. S5**). This was achieved by applying meshless Helmholtz-Hodge decomposition (Patane, 2022). Helmholtz-Hodge decomposition is commonly used to analyze vector fields in fluid dynamics, electromagnetism, and imaging (Bhatia *et al.*, 2013). The idea of Helmholtz-Hodge decomposition is to consider the vector field $f$ as a sum of a gradient vector field $g$ (i.e., curl-free) and a solenoidal vector field $r$ (i.e., divergence-free):

$$f(\boldsymbol{q}) = g(\boldsymbol{q}) + r(\boldsymbol{q}) = -\nabla\Phi(\boldsymbol{q}) + \nabla\times A(\boldsymbol{q}),$$

where $\boldsymbol{q}$ is the vector of independent variables (e.g., $\boldsymbol{q} = (x, \alpha, \beta)$ in our case), $\Phi(\boldsymbol{q})$ is a scalar potential, whose gradients are $g(\boldsymbol{q})$, and $A(\boldsymbol{q})$ is the vector potential, whose curl is $r(\boldsymbol{q})$. Practically, some numerical method first calculates the scalar potential $\Phi(\boldsymbol{q})$ by fitting a scalar function whose gradients best match the vector field (Hatton and Choset, 2011; Patane, 2022). This coincided with our goal. So, we applied Helmholtz-Hodge decomposition and used the scalar potential $\Phi(\boldsymbol{q})$ as the estimated potential energy landscape and its gradients $g(\boldsymbol{q})$ as the estimated landscape gradients.

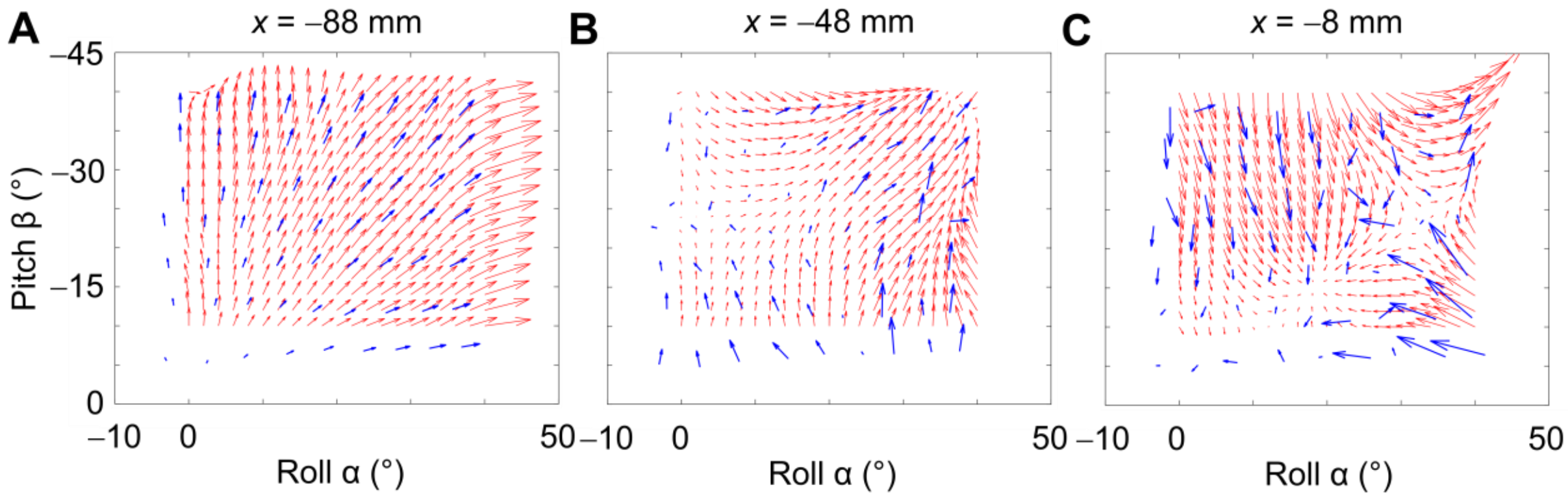

**Fig. S5. Meshless Helmholtz-Hodge decomposition results.** Representative input vector field (blue) and output potential energy landscape gradients (red), shown in the roll-pitch cross-section at the fore-aft position (**A**) $x$ = –88 mm, (**B**) $x$ = –48 mm, (**C**) $x$ = –8 mm.

Because the robot's body rotation changed slightly during each trial (Sec. 4.2), the measured robot coordinates were not strictly gridded (i.e., meshed) as required by most algorithms (Guo, Mandal and Li, 2005; Hatton and Choset, 2011; Lemoine *et al.*, 2015). We speculate that the inability to sample strictly on a grid in the state space also occurs in real-world robotic applications due to stochasticity arising from self-propulsion and frequent collisions with obstacles. To adapt to this, we proposed to perform a meshless algorithm (Razafindrazaka *et al.*, 2019; Patane, 2022).

The scalar potential function Φ and its gradients $g$ were approximated as a sum of the basis functions $\boldsymbol{\phi}$ (called kernel functions) and their gradients at $k$ scattered points (called centers):

$$\Phi = \sum_{i=1, \ldots, k} a_i \phi i = \boldsymbol{\phi}^{\mathrm{T}} \boldsymbol{a},$$

$$g = \sum_{i=1, \ldots, k} a_i \nabla\phi i = (\nabla\boldsymbol{\phi})^{\mathrm{T}} \boldsymbol{a},$$

where $\boldsymbol{a} = [a_1, \cdots, a_k]^{\mathrm{T}}$ are the coefficients of the linear combination. By matching the gradients with the vector field $\boldsymbol{V}$ of $n$ vectors using the least squares method, the coefficients are calculated as follows:

$$\boldsymbol{a} = [(\nabla\boldsymbol{\phi})^{\mathrm{T}}]^{-1} \boldsymbol{V}.$$

Because the Helmholtz-Hodge decomposition required the vector field to have similar ranges in all dimensions, we multiplied a ratio of 100 to the fore-aft force and a ratio of 0.01 = 1/100 to the fore-aft position ($x$) as inputs. We chose a commonly used Gaussian kernel function $\phi_i(\boldsymbol{q}) = \exp(-\sigma\|\boldsymbol{q} - \boldsymbol{q}_{c,1}\|)^2$, where $\boldsymbol{q}_{c,l}$ was the coordinate of the i-th center, because this kernel function was suitable for the expected continuous, 1-order smooth, non-periodic landscape. We generated 2000 centers by performing k-means clustering on the input coordinates. This kernel number yielded robust estimation results (from a preliminary test, it showed no significant performance reduction even with 60% data loss).

Note that the meshless Helmholtz-Hodge decomposition inherently induces errors in landscape estimation because it uses discrete data points. Even when performed on landscape gradients calculated from first principle, the landscape estimation had a relative error of 1% ± 3% in energy and 10% ± 23% in gradients.

## S.5 Estimating potential energy landscape from vision-based geometry

To assess how much force and torque sensing improved the accuracy of potential energy landscape estimation compared to vision-based geometry sensing, we generated a potential energy landscape where the beams were rigidly fixed, which can be estimated accurately from geometry-based sensing. Here, if we still assumed that the robot only moved in fore-aft, roll, and pitch directions (i.e., in the $x$-α-β space), we cannot appropriately define a finite potential energy when the robot had to penetrate a beam. Instead, the robot had to increase its vertical position $z$ to avoid penetrating the beams. The robot's gravitational potential energy at the minimum possible $z$ was then the system's potential energy.

## S.6 Oscillatory head flexion frequency selection

We selected the robot's oscillatory head flexion frequency based on the biological observation. When exploring and interacting with the beams, the cockroach flexed its head in an oscillatory manner with a variable frequency (**Fig. 2**). To calculate its average flexion frequency using data from (Wang, Othayoth and Li, 2022), we separated each head flexion cycle into two strokes—the downstroke and upstroke, where the head pitched downward and upward relative to the thorax (increasing and decreasing head flexion angle), respectively (**Fig. 2**). We defined the head flexion amplitude as the maximal range of head flexion angle. The head flexion amplitude averaged across all cycles in all trials observed was 15° ± 9° in the downstroke and 16° ± 10° in the upstroke. Furthermore, we calculated the angular velocity of oscillatory head flexion by dividing the amplitude by the duration of each stroke, to be 145° ± 100° $\cdot s^{-1}$ in the downstroke and 150° ± 90° $\cdot s^{-1}$ in the upstroke. Dividing the total amplitude (sum of upstroke and downstroke amplitudes) by the angular velocity (which was nearly the same for upstroke and downstroke), the average animal's oscillatory head flexion frequency was roughly 5 Hz. Because the animal traversed the beam obstacle in 4 ± 1 seconds from initial to final contact (Wang, Othayoth and Li, 2022), whereas the robot traversed in 10 seconds (**Sec. 4.2**), the robot's oscillatory head flexion frequency was chosen to be 2 Hz, so that both the animal and the robot's head flexed by 20 cycles over the entire traversal.

We speculated that the oscillatory head flexion only allowed the friction to cancel out at a high enough frequency, where it dominated the relative velocity. To test this, besides 2 Hz (results shown in **Fig. 7**), we also tried two other lower frequencies, 0.5 Hz and 1 Hz. We did not try a frequency higher than 2 Hz due to the speed limitations of the servo motors. All the data were processed by a filter with a cut-off frequency that was the same as the oscillatory head flexion (0.5 Hz or 1 Hz, respectively) (**Sec. S.9**). To test if oscillatory head flexion frequencies affected the match between the measured forces and torques with the conservative ones, we calculated the relative errors (see **Sec. S.9** for definitions) of the forces and torques in various oscillatory head flexion frequencies. We averaged these errors across the range of state space (fore-aft position, roll, pitch) tested. For both 0.5 Hz and 1 Hz, the head motion did not have a significant and substantial improvement in the match ($P > 0.05$ in Student's $t$-tests or Cohen's $|d| < 0.5$) in the edge contact case (**Fig. S6**). This supported our speculation.

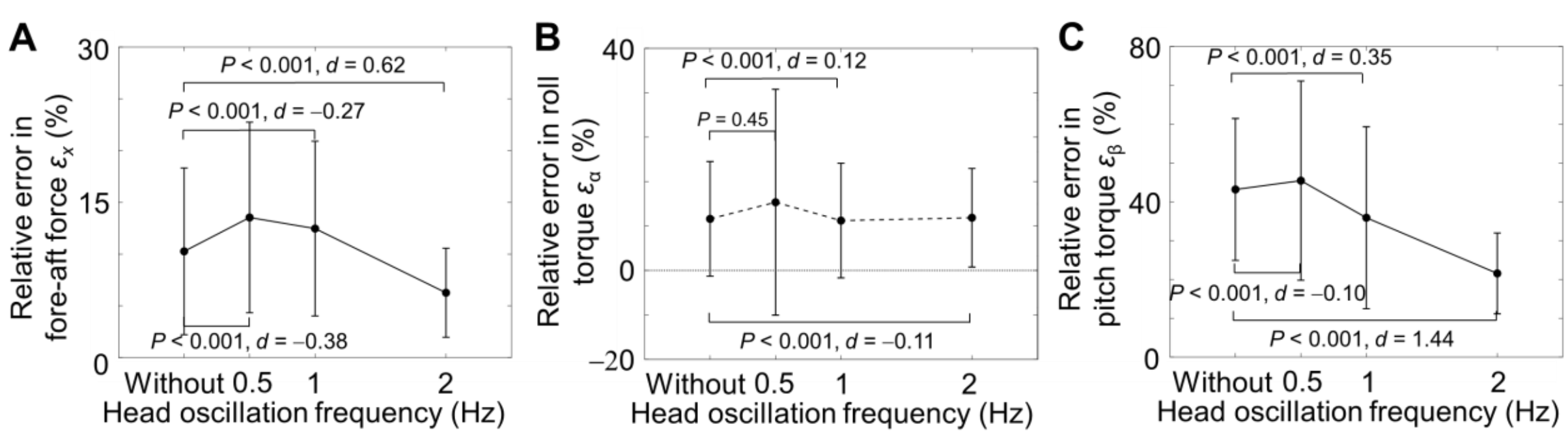

**Fig. S6. Effect of oscillatory head flexion frequency on the accuracy of potential energy landscape gradient estimation.** Relative error of **(A)** fore-aft force $F_x$ **(B)** roll torque $T_\alpha$, and **(C)** pitch torque $T_\beta$ and the conservative force and torques in edge contact phase without head flexion (0 Hz) and with head flexion at 0.5, 1, and 2 Hz, respectively. Points and error bars show means ± 1 standard deviation (s.d.) of data across the range of state space of (fore-aft position, roll, pitch) tested. *P* is the p-value from Student's *t*-test. *d* is the Cohen's *d*.

## S.7 Subtracting friction improved estimation of landscape gradients

To better estimate the landscape gradients in a face contact case, we tried subtracting friction and its torque from the measured ones. We assumed that the contact friction followed Coulomb's law of friction, which stated that the friction should be perpendicular to the normal direction of contact. To estimate the normal direction, we designed the pattern of contact sensors on the faces so that the variance of the normal direction on each sensor is small. The normal direction of each contact sensor was obtained as that at its center. With this, we achieved a highly accurate estimation (error < 5°, **Fig. 4J**, maximal normal direction error of each sensor was defined as the maximal angle between the normal directions at the corners and the center). Subtracting friction from the measured force should give the normal force (**Fig. 5A**, **B**, cyan), which was the projection of the measured force along the normal direction. The resulting torques were the cross-products of the position vector and the normal forces. The total normal forces and resulting torques were the sum of those from the two beams (**Fig. S7A-F**, red). Note that we did not define normal directions, normal forces, and their torques in the edge contact case.

To test if subtracting friction affected the match between the measured and normal forces and torques with the conservative ones, we calculated the relative errors (see **Sec. S.9** for definitions) of the forces and torques between without and with friction subtracted. We averaged these errors across the range of state space (fore-aft position, roll, pitch) tested.

In the face contact case, the normal fore-aft force better matched the conservative force than the raw measured force, with a significantly and substantially smaller relative error from $\varepsilon_x$ = 23% ± 7% to 4% ± 4% in the fore-aft ($x$) direction ($P < 0.001$, Student's $t$-test, Cohen's $d = 3.22$, **Fig. S7G**). On the other hand, the relative error did not differ significantly in the roll (α) direction ($\varepsilon_\alpha = 31\% \pm 70\%$ vs. 28% ± 71%, $P = 0.21$, Student's $t$-test, **Fig. S7H**), and differed significantly but not substantially in the pitch (β) direction ($\varepsilon_\beta = 20\% \pm 14\%$ vs. 22% ± 25%, $P < 0.001$, Student's $t$-test, Cohen's $d = -0.13$, **Fig. S7I**). These results showed that subtracting friction improved estimation of landscape gradients in one degree of freedom but not in the other two in our experiments. However, this method was less generalizable in other scenarios because, theoretically, it does not necessarily give conservative forces and torques.

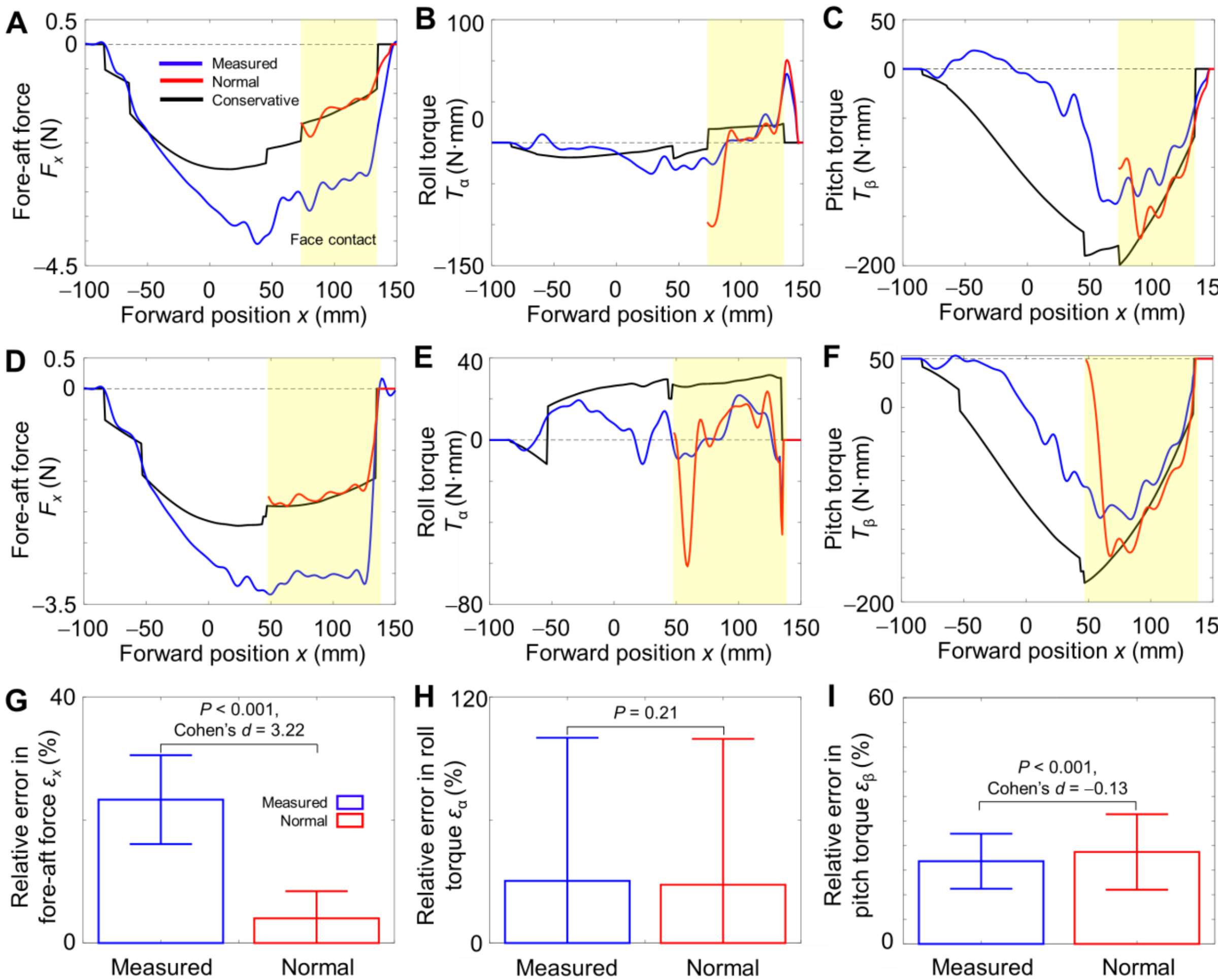

**Fig. S7. Subtracting friction improved estimation of landscape gradients.** (**A–F**) Measured and normal (A, D) fore-aft force $F_x$, (B, E) roll torque $T_\alpha$, and (C, F) pitch torque $T_\beta$, compared to the conservative force

and torques (i.e., from landscape), both as functions of fore-aft position $x$, from a representative trial. (A–C) body (roll, pitch) = (0°, −10°). (D–F) body (roll, pitch) = (15°, −20°). Blue, red, and black curves show measured, normal, and conservative data, respectively. Yellow shadow shows the range of $x$ position during which the interaction was in the face contact case. (**G–I**) Relative error ε of measured or normal force and torques. Blue and red show the measured and normal ones, respectively. Bars and error bars are means ± 1 standard deviation (s.d.) of data across the range of state space of (fore-aft position, roll, pitch) tested. $P$ is the p-value from Student's $t$-test. $d$ is the Cohen's $d$.

## S.8 Adaptation to a rounded robot body

To test how well our methods applied to a robot body without edges, we also made a rounded body of the same dimensions (**Fig. S8A, B**). We conducted the same experiment without oscillatory head flexion, measured contact forces and torques, and used them to estimate the potential energy landscape following similar steps as explained above.

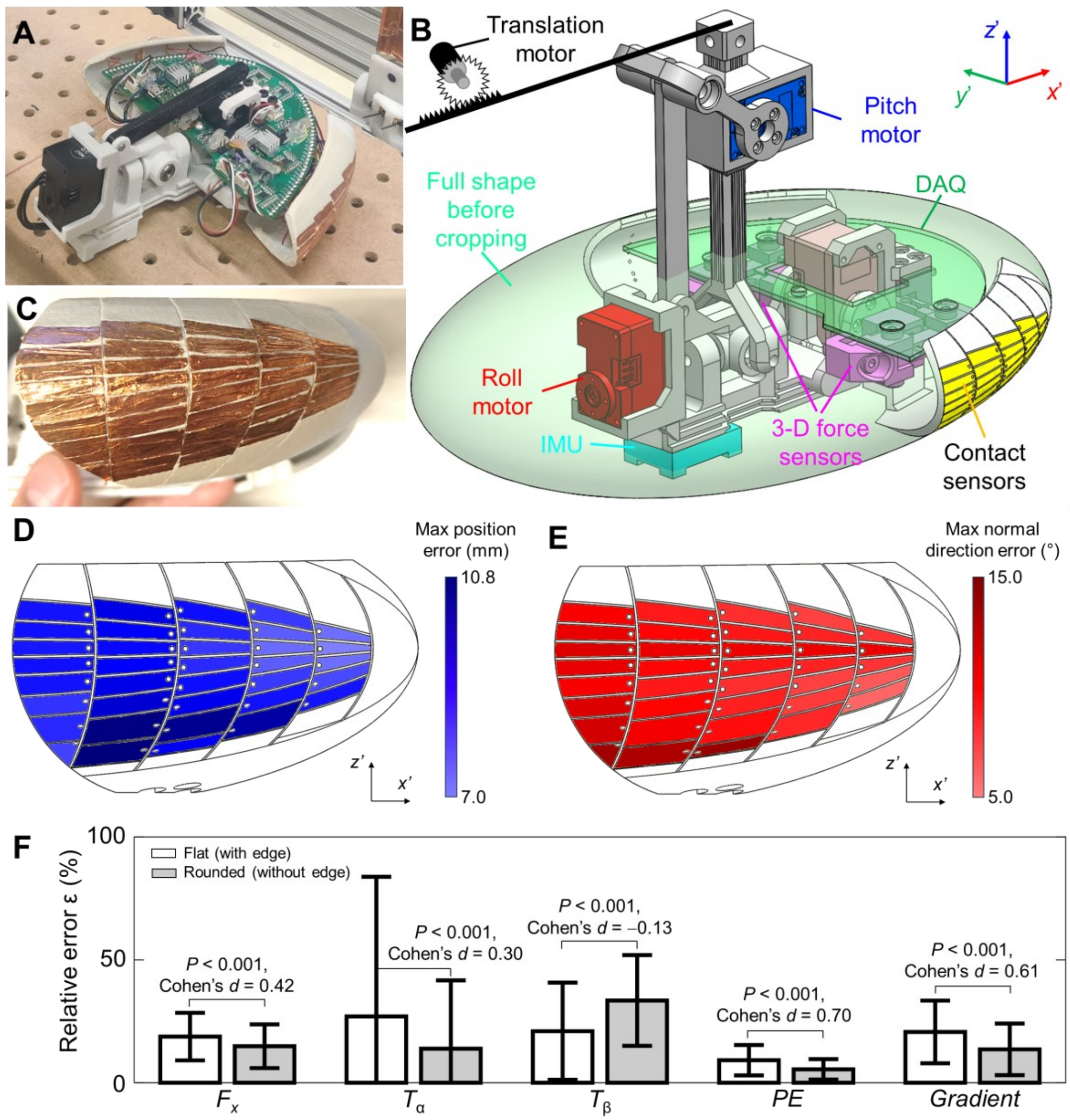

**Fig. S8. Experiments using a rounded body.** (**A**) Photo of the rounded body. (**B**) CAD model of the rounded robot. The shells were cropped from a perfect ellipsoid (translucent green) without edges. Other features were the same as the body in **Fig. 4B**. (**C**) Contact sensors from the side view. (**D, E**) Maximal (**D**) position and (**E**) normal direction errors from each contact sensor. Pattern shown is from the side view. *x*'- and *z*'-axes show body frame. Because the left and right shells are mirrored, only the right shell is shown. Color on each contact sensor shows error magnitude. (**F**) Relative error ε of contact force and torques and estimated landscape (*PE*) and its gradients. White and gray bars show using the body shape with a sharp edge in **Fig. 4A** and without sharp edges in A, respectively. In the first three groups, bars and error bars

show means ± 1 standard deviation (s.d.) of data across the range of state space of (fore-aft position, roll, pitch) tested. *P* is the p-value from Student's *t*-test. *d* is the Cohen's *d*.

The pattern of the contact sensors was carefully redesigned (**Fig. S8C**) to achieve high sensing accuracy in normal directions. Because the faces were highly curved, the normal direction changed dramatically around the equator (increasing by 107° over a 50 mm arc length). So, we prioritized minimizing the normal direction error (see **Sec. S.7** for definition). To maximize sensing area for a given threshold on normal direction error, the contact sensors were designed to be shortened along the latitude direction, roughly the same size, and placed contiguously without interstitial spaces. After multiple attempts, we increased the contact sensors to eight units in the latitude direction and decreased them to five in the longitude direction. Because the maximum number was restricted to 36 on each piece of shell due to the constraint from the DAQ, the number in the first two columns in the front was reduced. This design minimized the maximal normal direction errors (< 15°) (**Fig. S8D**) and gave low maximal position errors (< 11 mm) (**Fig. S8E**) over all the contact sensors.

To test if robot shape affected the match between the measured forces and torques with the conservative ones and the estimation of landscape and its gradients, we calculated the relative errors (see **Sec. S.9** for definitions) of forces, torques, potential energy landscapes, and gradients for the flat (with edge) and rounded (without edge) robot shape, respectively. We averaged these errors across the range of state space (fore-aft position, roll, pitch) tested.

With the rounded body, the measured contact forces and torques matched the conservative ones (i.e., from landscape gradients), with a reasonably small relative error of $\varepsilon_x$ = 15% ± 9% in the fore-aft ($x$) direction, $\varepsilon_\alpha$ = 14% ± 28% in the roll ($\alpha$) direction, and $\varepsilon_\beta$ = 33% ± 18% in the pitch ($\beta$) direction. The estimated landscape and its gradients matched those calculated from first principle, with a particularly small relative error of $\varepsilon_{PE}$ = 5% ± 4% in potential energy and a reasonably small relative error of $\varepsilon_{Grad}$ = 14% ± 11% in gradients. The change in shape had significant but not substantial effects in the matches in forces and torques ($P < 0.001$, Student's *t*-test, Cohen's $|d| < 0.5$) or even significantly and substantially improved the match in the potential energy landscape and its gradients ($P < 0.001$, Student's $t$-test, Cohen's $d > 0.5$) (**Fig. S8F**).

These results demonstrated that our contact force and torque sensing and potential energy landscape estimation methods also apply to a rounded robot body.

### S.9 Data processing and statistical tests

All the data (body fore-aft position $x$, roll $\alpha$, pitch $\beta$, head flexion angles, forces, torques as functions of time $t$) collected in experiments with oscillatory head flexion (**Sec. 2.5**, **Sec. S.6**) were processed by a six-order lowpass filter (i.e., Butterworth) with a cut-off frequency the same as the flexion (2 Hz in **Sec. 2.5**, 0.5 Hz or 1Hz in **Sec. S.6**) forwardly and backwardly (i.e., using "filtfilt" function in MATLAB) to avoid phase delay (this is similar to insects filtering out signals generated by their self-motion (Dickinson, 1990; Schmeling, Stange and Homberg, 2010; Kim, Fitzgerald and Maimon, 2015)). Because the data were highly consistent across repeated trials of the same experimental condition, we obtained the average data by linearly interpolating the measured data over the fore-aft position tested ($-100$ mm $\leq x \leq$ 150 mm) with an increment of 1 mm. Then, we averaged them over the five repeated trials. The average difference between measured and averaged data was 2% ± 1% in fore-aft force ($F_x$), 2% ± 2% for roll torque ($T_\alpha$), and 3% ± 3% for pitch torque ($T_\beta$) over the range of state space of (fore-aft position, roll, pitch) tested, confirming the consistency noted above.

To quantitatively compare measured forces and torques with the conservative ones, and the estimated potential energy landscape with that from first principle, we first determine the range of the fore-aft position ($x$) to compare in each trial. We only select fore-aft position $x$ from when the robot body first contacted both beams, determined as the smallest $x$ at which both the beam angles $\Delta\theta_1$ and $\Delta\theta_2$ exceeded a threshold of 1°, to the $x$ position when the robot body lost contact with both beams, defined as the smallest $x$ at which either beam angle $\Delta\theta_{1,2}$ reached maximum. We further identified the edge contact and face contact phase of this range of fore-aft positions ($x$) where the contacts with both beams are edge contacts or face contacts, respectively.

Note that we did not compare the forces and torques after the robot detached the beams (**Sec. 2.3**, **Sec. 2.5**), nor estimated the potential energy in the range of fore-aft position 100 mm < $x \leqslant$ 150 mm (**Sec. 2.4**), because the potential energy model adapted from (Othayoth, Thoms and Li, 2020) assumed the robot body to be a full shape, whereas our robot only had a head at the front (**Sec. 2.2**), the contact forces and torques should not approximate the landscape gradients in these ranges.

When comparing the estimated landscape gradients with those from modeling, we rescaled the gradients in the roll and pitch degrees of freedom to be comparable with those in the fore-aft translation axis. The rescaled gradient had the same value in the fore-aft translation ($x$) axis, expressed in units of newtons (N), and was 1/100 of the value used in the roll (α) and pitch (β) degrees of freedom, which were expressed in newton-millimeters (N·mm).

To quantify the match between the measured forces and torques and the conservative ones, as well as the estimation accuracy of potential energy landscape and its gradients, we defined the relative error ε as the difference (absolute value) between the measured or estimated data $d$ and the data from first principle $r$, normalized by the characteristic scale $CS$ along the relevant degree of freedom:

$$\varepsilon = |d - r|/CS \times 100\%.$$

We then averaged the relative error over the state space of (fore-aft position, roll, pitch) tested. The characteristic scales of forces and torques were the maximum measured value (absolute value) in the experiments without head oscillation ($F_{x,CS}$ = 4.48 N, $T_{\alpha,CS}$ = 137 N·mm, $T_{\beta,CS}$ = 150 N·mm). The characteristic scale of potential energy landscapes is the range of the potential energy landscape from first principle ($PE_{CS}$ = 337 mJ). The characteristic scale of landscape gradients is the maximum magnitude of the landscape gradients from first principle after rescaling (see previous paragraph) ($Grad_{CS}$ = 413 units).

To test if oscillatory head flexion affected the match between the measured forces and torques with the conservative ones (**Sec. 3.4, Sec 5.6**), we calculated the relative errors of forces and torques without and with oscillatory head flexion, and we averaged them over the range of state space (fore-aft position, roll, pitch) for edge contacts and face contacts, respectively.

To test if subtracting friction affected the match between the forces and torques with the conservative ones (**Sec 5.7**), we calculated the relative errors of forces and torques without and with friction subtracted, and we averaged them over the range of state space (fore-aft position, roll, pitch) for face contacts.

To test if the robot body shape affected the match between the forces and torques with the conservative ones, and the match between the estimated potential energy landscape and gradients with that from first principle (**Sec 5.8**), we calculated the relative errors of forces and torques, potential energy landscape, and gradients with the two body shapes, respectively, and we averaged them over the range of state space (fore-aft position, roll, pitch).

We used Student's $t$-tests to test whether the treatments above had a significant effect. When a significant effect is detected, we further use Cohen's $d$ to assess how strong the effect was, where Cohen's $d > 0.5$ indicates a substantial decreasing effect, $d < -0.5$ indicates a substantial increasing effect, and $|d| > 0.5$ indicates a minor effect. A treatment had a strong benefit for estimation only when it significantly ($P < 0.05$) and substantially (Cohen's $d > 0.5$) reduced the relative error.

All average data were reported as mean ± 1 standard deviation (s.d.). All data analyses and statistical tests were performed using MATLAB R2021b (MathWorks, MA).

**Video 1. Experiment recording and data acquisition.** Video recording of experiments (left) and real-time data visualization (right). (**Part 1**) Without oscillatory head flexion. (**Part 2**) With oscillatory head flexion.

**Video 2. Evolution of the estimated potential energy landscape.** The estimated potential energy landscape (top, left) changed as the body moved forward, compared with that calculated from first principle (right). The video recording (bottom, left) is from a representative trial, similar to those shown in **Figs. 5C, E**.

**Video 3. Relative errors.** The relative errors of (**Part 1**) fore-aft force (top, left), roll torque (top, right), pitch torque (bottom, right), (**Part 2**) potential energy landscape (top, left), and landscape gradients (top, right), as the body moved forward. Data are from without oscillatory head flexion. Forces and torques data (Part 1, bottom, left) and video recording (Part 2, bottom, left) are from a representative trial, the same as in **Fig. 5C, E**. Note that the color bar for the relative error of the potential energy landscape is capped at 10%, not 100% as for other data, for better visualization.